\documentclass[10pt,journal,compsoc]{IEEEtran}
\usepackage{url}  
\usepackage[colorlinks, linkcolor=blue, anchorcolor=black, citecolor=blue]{hyperref}
\usepackage{epsfig}
\usepackage{graphicx}
\usepackage{bbding} 
\usepackage{pifont} 
\usepackage{wrapfig} 
\usepackage{microtype}
\usepackage{amsfonts}
\usepackage{bm}
\usepackage{soul}
\usepackage{amsmath}
\usepackage{booktabs}
\usepackage{amsmath,amssymb}
\newtheorem{theorem}{Theorem}
\newtheorem{lemma}{Lemma}
\newtheorem{principle}{Principle}

\usepackage{mathrsfs} 
\usepackage{algorithmic,algorithm} 
\usepackage{multirow} 
\usepackage{subfigure} 
\usepackage{bbding} 
\usepackage{pifont} 
\usepackage{wrapfig}  
\usepackage{microtype} 
\usepackage{amsfonts} 
\usepackage{array,caption} 
\usepackage{bbm}
\usepackage[dvipsnames]{xcolor,colortbl} 
\usepackage{soul}
\usepackage{microtype} 

\usepackage[normalem]{ulem} % use normalem to protect \emph
\usepackage{mdframed}
\usepackage{enumitem}

\newcommand{\cmark}{\ding{51}}
\newcommand{\xmark}{\ding{55}}

\newcommand{\changed}[1]{\textcolor{black}{#1}}
\newcommand{\changedxx}[1]{\textcolor{black}{#1}}

\newcommand{\changedrr}[1]{\textcolor{black}{#1}} %varcolor
\newcommand{\changedrrr}[1]{\textcolor{black}{#1}} %varcolor

\def\modelnameshort{CausalDA}  
\def\modelname{latent Causal factors discovery}

\definecolor{cmblu}{RGB}{51,102,240} 
\definecolor{cmred}{RGB}{241,22,22} 

\definecolor{varcolor}{RGB}{0,153,102}

\ifCLASSOPTIONcompsoc
  \usepackage[nocompress]{cite}
\else
  \usepackage{cite}
\fi
\ifCLASSINFOpdf
\else
\fi
\begin{document}
%
% paper title
% Titles are generally capitalized except for words such as a, an, and, as,
% at, but, by, for, in, nor, of, on, or, the, to and up, which are usually
% not capitalized unless they are the first or last word of the title.
% Linebreaks \\ can be used within to get better formatting as desired.
% Do not put math or special symbols in the title.
\title{Unified Source-Free Domain Adaptation} 
% Knowledge

%
%
% author names and IEEE memberships
% note positions of commas and nonbreaking spaces ( ~ ) LaTeX will not break
% a structure at a ~ so this keeps an author's name from being broken across
% two lines.
% use \thanks{} to gain access to the first footnote area
% a separate \thanks must be used for each paragraph as LaTeX2e's \thanks
% was not built to handle multiple paragraphs
% Changshui~Zhang,~\IEEEmembership{Fellow,~IEEE,}
%

\author{Song~Tang,
        Wenxin~Su,
        Mao~Ye{\Envelope},
        % Jianwei~Zhang,
        Boyu~Wang,
        and~Xiatian~Zhu{\Envelope}
        
\thanks{\textit{{\Envelope} Corresponding authors: Mao Ye, and Xiatian Zhu.}} 
\thanks{Song~Tang is with the Institute of Machine Intelligence, University of Shanghai for Science and Technology, Shanghai, China; the TAMS Group, Department of Informatics, Universität Hamburg, Hamburg, Germany.}
\thanks{Wenxin~Su is with the Institute of Machine Intelligence, University of Shanghai for Science and Technology, Shanghai, China.}
\thanks{Mao~Ye is with the School of Computer Science and Engineering, University of Electronic Science and Technology of China, Chengdu, China.}
% \thanks{Jianwei~Zhang is with the TAMS Group, Department of Informatics, Universität Hamburg, Hamburg, Germany.}
\thanks{Boyu~Wang is is with the Vector Institute, the Department of Computer Science, University of Western Ontario, London, Canada.}
\thanks{Xiatian~Zhu is with the Surrey Institute for People-Centred Artificial Intelligence, and Centre for Vision, Speech and Signal Processing, University of Surrey, Guildford, UK.}

\thanks{Manuscript received April 19, 20**; revised August 26, 20**.}
}

% note the % following the last \IEEEmembership and also \thanks - 
% these prevent an unwanted space from occurring between the last author name
% and the end of the author line. i.e., if you had this:
% 
% \author{....lastname \thanks{...} \thanks{...} }
%                     ^------------^------------^----Do not want these spaces!
%
% a space would be appended to the last name and could cause every name on that
% line to be shifted left slightly. This is one of those "LaTeX things". For
% instance, "\textbf{A} \textbf{B}" will typeset as "A B" not "AB". To get
% "AB" then you have to do: "\textbf{A}\textbf{B}"
% \thanks is no different in this regard, so shield the last } of each \thanks
% that ends a line with a % and do not let a space in before the next \thanks.
% Spaces after \IEEEmembership other than the last one are OK (and needed) as
% you are supposed to have spaces between the names. For what it is worth,
% this is a minor point as most people would not even notice if the said evil
% space somehow managed to creep in.

% The paper headers
\markboth{SUBMITTED TO IEEE TRANSACTIONS ON PATTERN ANALYSIS AND MACHINE INTELLIGENCE}%
{Shell \MakeLowercase{\textit{et al.}}: Bare Demo of IEEEtran.cls for ** Journals}
% The only time the second header will appear is for the odd numbered pages
% after the title page when using the twoside option.
% 
% *** Note that you probably will NOT want to include the author's ***
% *** name in the headers of peer review papers.                   ***
% You can use \ifCLASSOPTIONpeerreview for conditional compilation here if
% you desire.

% The publisher's ID mark at the bottom of the page is less important with
% Computer Society journal papers as those publications place the marks
% outside of the main text columns and, therefore, unlike regular IEEE
% journals, the available text space is not reduced by their presence.
% If you want to put a publisher's ID mark on the page you can do it like
% this:
%\IEEEpubid{0000--0000/00\$00.00~\copyright~2015 IEEE}
% or like this to get the Computer Society new two part style.
%\IEEEpubid{\makebox[\columnwidth]{\hfill 0000--0000/00/\$00.00~\copyright~2015 IEEE}%
%\hspace{\columnsep}\makebox[\columnwidth]{Published by the IEEE Computer Society\hfill}}
% Remember, if you use this you must call \IEEEpubidadjcol in the second
% column for its text to clear the IEEEpubid mark (Computer Society jorunal
% papers don't need this extra clearance.)

% use for special paper notices
%\IEEEspecialpapernotice{(Invited Paper)}

% for Computer Society papers, we must declare the abstract and index terms
% PRIOR to the title within the \IEEEtitleabstractindextext IEEEtran
% command as these need to go into the title area created by \maketitle.
% As a general rule, do not put math, special symbols or citations
% in the abstract or keywords.
\IEEEtitleabstractindextext{%

\begin{abstract}
In the pursuit of transferring a source model to a target domain without access to the source training data, Source-Free Domain Adaptation (SFDA) has been extensively explored across various scenarios, including Closed-set, Open-set, Partial-set, and Generalized settings. Existing methods, focusing on specific scenarios, not only address \changedrr{a limited subset} of challenges but also necessitate prior knowledge of the target domain, significantly limiting their practical utility and deployability.
In light of these considerations, we introduce a more practical yet challenging problem, termed {\it unified SFDA}, which comprehensively incorporates all specific scenarios in a unified manner. 
\changed{In this paper, we propose a novel approach {\it {\modelname} for unified SFDA} (\modelnameshort).} 
In contrast to previous alternatives that emphasize learning the statistical description of reality, we formulate {\modelnameshort} from a causality perspective. The objective is to uncover \changedrrr{potential causality} between latent variables and model decisions, enhancing the reliability and robustness of the learned model against domain shifts.  
To integrate extensive world knowledge, we leverage a pre-trained vision-language model such as CLIP. This aids in the formation and discovery of latent causal factors in the absence of supervision in the variation of distribution and semantics, coupled with a newly designed information bottleneck with theoretical guarantees. Extensive experiments demonstrate that {\modelnameshort} can achieve new state-of-the-art results in distinct SFDA settings, as well as source-free out-of-distribution generalization.
Our code and data are available at \url{https://github.com/tntek/CausalDA}.
\end{abstract}

}
% make the title area
\maketitle
% To allow for easy dual compilation without having to reenter the
% abstract/keywords data, the \IEEEtitleabstractindextext text will
% not be used in maketitle, but will appear (i.e., to be "transported")
% here as \IEEEdisplaynontitleabstractindextext when the compsoc 
% or transmag modes are not selected <OR> if conference mode is selected 
% - because all conference papers position the abstract like regular  
% papers do.
\IEEEdisplaynontitleabstractindextext  
% \IEEEdisplaynontitleabstractindextext has no effect when using
% compsoc or transmag under a non-conference mode.

% For peer review papers, you can put extra information on the cover
% page as needed:
% \ifCLASSOPTIONpeerreview
% \begin{center} \bfseries EDICS Category: 3-BBND \end{center}
% \fi
%
% For peerreview papers, this IEEEtran command inserts a page break and
% creates the second title. It will be ignored for other modes.
\IEEEpeerreviewmaketitle

\section{Introduction} \label{sec:introduction}

\IEEEPARstart{D}{ue} to   demands for privacy and information protection in public security and commercial competition, well-annotated training data are recognized as crucial for varying purposes, requiring stringent access control. However, in practical scenarios, the use of a source model (pre-trained on the source domain) without access to the actual source data, referred to as Source-Free Domain Adaptation (SFDA) \cite{kim2021domain}, has become more feasible. This approach involves the challenges posed by strict data access controls and emphasizes the adaptation of a pre-trained model to new domains without relying on the availability of the original source data. 

SFDA has been explored across various scenarios, as summarized in Tab.~\ref{tab:sfda-comp}. The most constrained scenario, termed {\it Closed-set}, assumes identical categories of interest in both the source and target domains, aiming to address covariate shift, such as domain-specific appearance distributions \cite{li2020model}. {\it Generalized SFDA} extends this by requiring the adapted model to perform well in both the source and target domains without forgetting the source domain \cite{yang2021generalized}. Moving beyond the vanilla category identity constraint, {\it Open-set} SFDA \cite{panareda2017open} considers the target domain with additional categories, while {\it Partial-set} SFDA \cite{cao2019learning} takes the opposite direction. Both scenarios require handling the additional challenge of semantic shift.

Typically, prevalent SFDA methods tend to concentrate on particular scenarios, tackling only a subset of challenges. For instance, SFDA-DE\cite{ding2022source} focuses on Closed-set scenarios, SF-PGL\cite{luo2023source} on Open-set scenarios, PSAT\cite{tang2023psat} on Generalized settings, and CRS \cite{zhang2023crs} on Partial-set scenarios. 
Nevertheless, this approach significantly restricts their practical utility and deployability, given that we often possess limited prior knowledge about the target domain and minimal control or selection over its conditions.

\begin{table}[t]
    \caption{Summary of different SFDA settings. $C_s$/$C_t$: The category set of the source/target domain.}
    \label{tab:sfda-comp}
    \renewcommand\tabcolsep{3.0pt}
    \renewcommand\arraystretch{1.0}
    % \footnotesize
    % \small
    \scriptsize
    \centering
    \begin{tabular}{ l | c c c c }
        \toprule
        Setting       & Covariate shift  &Semantic shift &Anti-forgetting & Category \\
        \midrule
        Closed-set    &\cmark  &\xmark  &\xmark  &$C_s = C_t$        \\
        Generalized   &\cmark  &\xmark  &\cmark  &$C_s = C_t$        \\
        Open-set      &\cmark  &\cmark  &\xmark  &$C_s \subset C_t$  \\
        Partial-set   &\cmark  &\cmark  &\xmark  &$C_s \supset C_t$  \\ \hline
        \bf Unified (Ours)  &\cmark  &\cmark  &\cmark & Any \\
        \bottomrule
    \end{tabular}
\end{table}

To address the aforementioned limitation, this study introduces 
a more realistic yet challenging problem, referred to as {\bf\it Unified Source-Free Domain Adaptation} ({\it Unified SFDA}), with the goal of comprehensively addressing all the specific scenarios mentioned in a unified manner. To achieve this, \changed{we propose a novel approach {\it latent Causal discovery of salient factors for unified SFDA (\modelnameshort)},} going beyond conventional statistical association learning of related variables by exploring the underlying causal relationships \cite{pearl2009causality}. The causal mechanism discovered is expected to be more reliable in varying distributions and semantic contexts, providing a unified solution for different scenarios of SFDA.

In the absence of label supervision for both distributional and semantic variations, our {\modelnameshort} is specifically formulated using a structural causal model in the logits space. The latent causal factors are disentangled into two complementary parts: 
(i) {\it external causal factors} and (ii) {\it internal causal factors, consider that the information of both domains is not necessarily complete.} 
For the former, we leverage a pre-trained large Vision-Language (ViL) model with rich knowledge such as CLIP~\cite{radford2021learning}, which has been exposed to a vast amount of multimodality information sources. 
The latter is identified under the guidance of discovered external causal factors. 
The discovery of external and internal causal factors is alternated by designing a self-supervised information bottleneck with theoretical guarantees.

Our {\bf contributions} are summarized as follows:
\begin{itemize}
    \item [(i)] For the first time, we introduce a unified SFDA problem that comprehensively incorporates various specific scenarios. This design eliminates the need for prior knowledge about the target domain before model deployment, enhancing practical usability and applicability in real-world scenarios.
    
    \item [(ii)]  We propose a novel {\modelnameshort} approach for unified SFDA.
    Instead of learning the statistical description of problem reality as conventional methods do, {\modelnameshort} is formulated under a causality perspective. 
    It \changedrrr{attempts} to reveal the potential causality between latent variables and model decisions, providing favorable robustness against both distributional and semantic shifts.
    
    \item [(iii)] Extensive experiments demonstrate that {\modelnameshort} consistently achieves new state-of-the-art results on varying SFDA scenarios as well as source-free out-of-distribution generalization.  
\end{itemize}

\section{Related Work}\label{sec:rework}

\subsection{Source-free domain adaptation}

Most of the prior SFDA methods consider the Closed-set setting where the source and target domains share the same classes, and the focus is \changedrr{on} cross-domain distribution alignment.
Existing methods exist in two categories.  
The first converts SFDA to Unsupervised Domain Adaptation~(UDA) by constructing a pseudo-source domain~\cite{2020MA}, exploiting source prototype-guided data generation~\cite{tian2022vdm} or splitting source-like subset from target data~\cite{du2021ps}. 
The second follows the paradigm of self-supervised learning~\cite{liang2020we,chen2022self}, introducing self-guidance.
Besides widely used pseudo-labels~\cite{liang2020we, litrico2023guiding}, geometry information~\cite{yang2021nrc,yang2023trust,tang2022sclm}, \changedrr{contrastive} data~\cite{zhang2023crs} and historical data~\cite{tang2023psat} have also been exploited.

Recently, more practical settings have been studied.
For example, the Generalized setting shifts aims to mitigate forgetting the source domain \cite{yang2021generalized}.
A representative approach is combining \changedrr{continual} learning techniques with cross-domain adaptation, e.g., the domain attention-based gradient regulation~\cite{yang2021generalized} and source guidance constructed by historical information refining~\cite{tang2023psat}.
There exist works considering Open-set \cite{luo2023source} and Partial-set \cite{li2022partial} settings. 
For instance, to control the Open-set risk, a progressive balanced pseudo-labeling strategy \cite{luo2023source}, and Closed-set class prototypes are proposed~\cite{vray2024distill}. 
Estimating the target data distribution helps with negative transfer reduction from the partial classes shift \cite{lee2022confidence}.

Alternatively, some works were proposed to enhance the discriminating ability for out-of-distribution classes \cite{samadh2023align}. There are two main strategies: Expanding inter-class distance to prevent semantic confusion~\cite{tang2021model,liang2020we},
and introducing external semantics, e.g., CLIP, to match out-of-distribution~\cite{shu2022test}.

Most recently, multimodal large models such as CLIP and ShareGPT~\cite{chen2024sharegpt4v} have been introduced to advance SFDA, substantially enhancing adaptation performance by leveraging generic knowledge in these models.
Two strategies are adopted: (1) customizing and adapting this knowledge via mutual knowledge distillation~\cite{tang2024source,zhang2025source} or multi-teacher guidance~\cite{chen2024empowering} or multimodal space alignment~\cite{chen2025source}; and (2) further denoising ViL predictions \cite{tang2025proxy}.
Critically, we highlight a couple of key aspects that make {\modelnameshort} conceptually distinct from these SFDA methods.
First, CausalDA is unique in design philosophy by learning a causal representation, in contrast to the statistical knowledge DIFO \cite{tang2024source} and ProDe \cite{tang2025proxy} both are designed to learn.
As a consequence, the outreach and generalization can be further extended, leading to the first dedicated unified SFDA approach.
Instead, DIFO and ProDe predominantly focus on the conventional closed-set setting.
Further, this conceptual discrepancy leads to distinct leveraging of ViL knowledge. 
DIFO and ProDe both employ the generic ViL knowledge directly as supervision, fostering the establishment of statistical relationships, whilst {\modelnameshort} utilizes ViL knowledge to uncover the underlying causal factors for capturing the essence of visual recognition.

Existing work targets only specific SFDA settings, limiting its practical applicability and generality. 
This paper introduce a unified SFDA setting that incorporates all the pre-existing scenarios.
With this new setup, we aim to foster more advanced SFDA methods that could generalize to a variety of problem settings without the need for designing and maintaining an array of distinct algorithms. 

% Existing work as above focuses on a specific SFDA setting, substantially limiting its usability and generality in practice. 
% To address this limitation, we introduce a unified SFDA setting that incorporates all the pre-existing scenarios.
% With this new setup, we aim to foster more advanced SFDA methods that could generalize to a variety of problem settings without the need for designing and maintaining an array of distinct algorithms. 

% by modeling potential statistical correlations

% \changedrr{Notably, although the ViL-based methods demonstrate remarkable performance in adapting across multiple SFDA settings due to the integration of generic ViL knowledge, they were originally designed for the vanilla Closed-set setting.} 

\subsection{Causality methods in transfer learning} \label{sec:rela-causal}
Causality methods aim for a robust function relation between random variables, which is insensitive to the extra variation/disturbing \cite{scholkopf2021toward}. 
This property leads to recent popularity of this new paradigm in transfer learning, e.g., UDA~\cite{wu2024causality}, domain generalization (DG)~\cite{christiansen2021causal}, and out of distribution (OOD)~\cite{byun2025ccl}.
% wang2022causal
These attempts aim to reduce non-causal factors by introducing artificial intervention.
For example, the probability distribution change (used to represent the domain shift)  and the semantic consistency constraint are jointly employed to implement this intervention.  
As source labels are available to ensure consistency, \changedrr{these methods emphasize constructing the distribution change in two lines.} 
One explicitly generates augmented data to disturb the distribution, e.g., the non-linear augmentation and spatially-variable blending augmentation~\cite{ouyang2022causality}, inverse Fourier transformation augmentation~\cite{lv2022causality}, generative models-based cross-domain image style transformation~\cite{wang2022out}, and causality-adjusted augmentation~\cite{zhang2025causality}. 
The other implicitly modifies the distribution by exploiting cross-domain data. 
For example, MatchDG~\cite{mahajan2021impinter} developed cross-domain contrastive learning where the input's positive sample was randomly selected from samples with the same class. 
COR~\cite{wang2022causal} adopted the variational inference encoder to infer unobserved causal factors from historical data, taking the time domain as a natural distribution change.

Compared with the existing works, the following three features distinguish {\modelnameshort} from them.
(i) {\modelnameshort} does not rely on real labels that are indispensable for previous works. 
(ii) {\modelnameshort} discovers the causal factors instead of indirectly removing the non-causal ones. 
(iii) {\modelnameshort} builds the structural causal model at the logits level, whilst previous methods construct on raw data.

\begin{figure}[t]
    \begin{center}
        \includegraphics[width=0.8\linewidth]{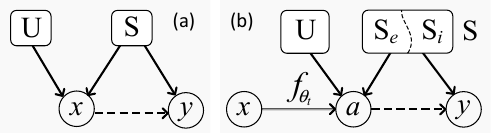}
    \end{center}
    \caption{
    Comparison of Structural Causal Models (SCM). 
    {\bf (a)} SCM for transfer learning with source domain,
    where $S$ and $U$ represent causal and non-causal factors. Domain shift is generally caused by $U$,
    which, together with the generalizable information $S$, e.g.,
    % such as (b) 
    the shape/structure of a dog (see Fig.~\ref{fig:domain-shift}), 
    form the observation $x$. 
    {\bf (b)} Our proposed SCM for SFDA in a latent space (e.g., the logit $\boldsymbol{a}$)
    where the causal factors $S$ are decomposed into the external $S_e$ and internal $S_{i}$ components.   
    }
    \label{fig:scm}
\end{figure}

\section{Methodology} \label{sec:method}

\subsection{Unified SFDA problem}
SFDA aims to transfer a model pre-trained on the source domain to a different but related target domain without labeling. Formally, let $\mathcal{X}_s\!=\!\{\boldsymbol{x}_{i}^s\}_{i=1}^{n_s}$ and $\mathcal{Y}_s\!=\!\{{y}_{i}^s\}_{i=1}^{n_s}$ be the source samples and their truth labels.
Similarly, the unlabeled samples and the truth target labels are $\mathcal{X}\!=\!\{\boldsymbol{x}_{i}\}_{i=1}^{n}$ and $\mathcal{Y}\!=\!\{{y}_{i}\}_{i=1}^{n}$, respectively. 
The unified SFDA task is to learn a target model $f_{\theta_t}:\mathcal{X} \to \mathcal{Y}$ {\it without any prior knowledge of the target domain not the relationships between the source and target domains (see Tab.~{\ref{tab:sfda-comp}})}, given (i) the source model $f_{\theta_s}$ pre-trained over $\mathcal{X}_s$, $\mathcal{Y}_s$, and (ii) the unlabeled target data $\mathcal{X}$.

\subsection{Causality in transfer learning with source domain} \label{sec:causal-dg-uda}
\begin{figure}[t]
    \setlength{\abovecaptionskip}{0cm}
    \begin{center}
        \subfigure[]{
            \includegraphics[width=0.45\linewidth,height=0.27\linewidth]{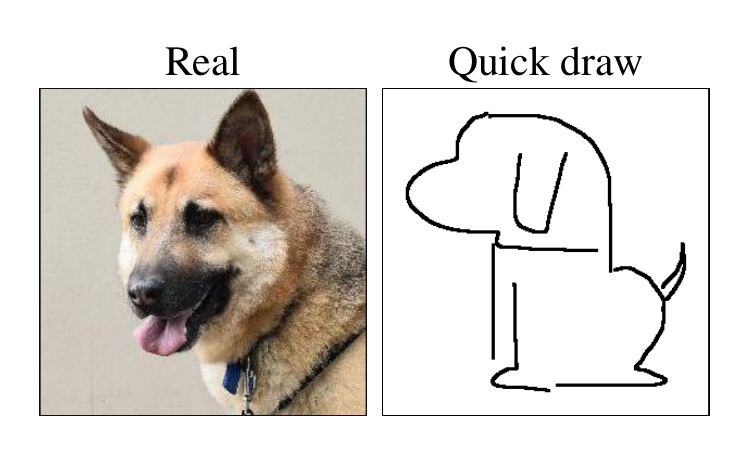}}
        \subfigure[]{
            \includegraphics[width=0.45\linewidth,height=0.27\linewidth]{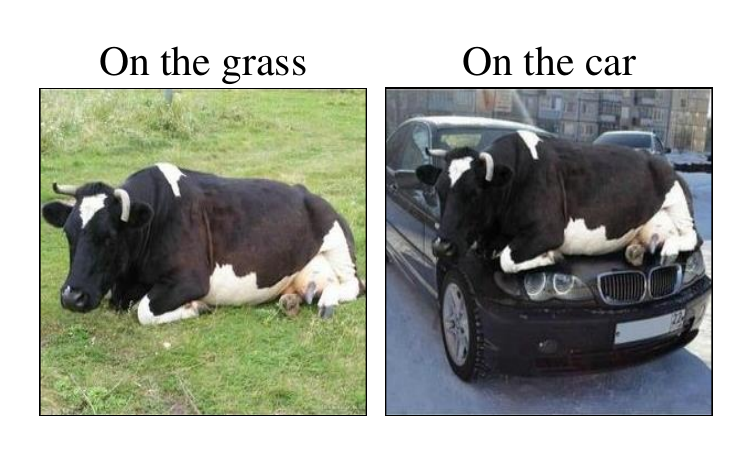}}
    \end{center}
    \setlength{\abovecaptionskip}{0cm}
    \caption{  
    Illustration of domain shift. 
    (a)~{\bf Different appearance styles} of dog;
    (b)~{\bf Spurious dependence of background:} Cow in grass ground vs. on a car.   
    }
    \label{fig:domain-shift}
\end{figure}

In transfer learning \changedrr{with available source domain}, such as DG and UDA, the causality relation hidden in the statistical dependence (between raw image $\boldsymbol{x}$ and available label $\boldsymbol{y}$) can be summarized to a Structural Causal Model (SCM). 
As depicted in Fig.~\ref{fig:scm}(a), $\boldsymbol{x}$ is generated jointly by non-causal factors ${U}$ and causal factors ${S}$, while label $\boldsymbol{y}$ is determined by ${S}$ alone. 
Here, non-causal factors collectively represent the latent variables that generate the category-independent appearance of images, e.g., different domain styles and spurious dependence (Fig.~\ref{fig:domain-shift}); The causal factors present the ones determining the classification, e.g., the object shape.

In a causal view, solving the transfer learning involves eliminating the domain shift by building a robust classification function, i.e., $P(\boldsymbol{y}|S)$. 
With SCM, we need to disconnect $U$ to $\boldsymbol{x}$.  
So existing works impose intervention upon $\boldsymbol{x}$, according to the intervention theory~\cite{pearl2009causality}. 
Formally, this scheme can be formulated as 
\begin{equation}
    \label{eqn:intervention}
    \begin{split}
        P(\boldsymbol{y}|S) = P(\boldsymbol{y}|do(x), S), 
    \end{split}
\end{equation}
where $do(\cdot)$ means the do-operation of imposing intervention upon variables.
% , which is implemented by consistency learning with distribution disturbing. 

In summary, the key idea \changedrr{is that when} we change the probability distribution of non-causal factors (intervention) whilst keeping the category (prediction consistency), the causality can be extracted. 
Often, domain-related data augmentation is adopted as an intervention, whilst the source labels ensure prediction consistency.
However, both conditions are unavailable in SFDA, making it inapplicable.
% cross-domain data are employed to construct domain-shift-presentation-centric augmentation as an intervention, 

\subsection{Latent causality in unified SFDA}
To address the challenges outlined earlier, we propose an approach named \textit{\modelname} ({\modelnameshort}), taking into account the following considerations. Instead of extracting causality in the raw input space $\boldsymbol{x}$, we shift our focus to the logit space ${\bm{A}}$: $\boldsymbol{a}=f_{\theta_t}(\boldsymbol{x}) \in \bm{A}$. 
The choice of this space for causality analysis is crucial for two main reasons:
\changed{
First and foremost, from a causal perspective, using a post-intervened space enhances causality capture by incorporating a probability distribution shift into the data \cite{mahajan2021impinter,wang2022causal,pearl2009causality}.
In our case, the logit space is post-intervened due to the inherent distribution shift between the in-training target model (initialized with the pre-trained source domain model) and the target domain. Specifically, the target model’s logits for the target domain data reflect this domain shift, making them a particular form of post-intervention in this context.
}

\changed{
Furthermore, the logit space is both highly semantic and compact, making it more efficient to manipulate. Since logits encode rich semantic information about class relationships, operating in this space allows for better generalization and transferability across domains. Additionally, its compact representation reduces computational complexity, facilitating more efficient learning and adaptation. These properties collectively make the logit space a more effective choice for our approach.  
}
\changedrr{Tab.~\ref{tab:connection-our-conven} presents an analogy of {\modelnameshort} and the existing causal learning approach.}

\begin{table}[t]
    \renewcommand\tabcolsep{10pt}
    \renewcommand\arraystretch{1.2}
    \scriptsize
    \centering
    \begin{tabular}{l|c c}
        \toprule
        Approach & Intervention & Consistency \\
        \midrule
        Existing~\cite{chen2021domain,christiansen2021causal,wang2022causal} & Data augmentation     & Supervised \\
        {\bf \modelnameshort}          & Using logit  & Unsupervised \\
        \bottomrule
    \end{tabular}
    \caption{
    \changedrr{Analogy of causality learning.}
    %The design connection between {\modelnameshort} and %conventional causal scheme
    }
    \label{tab:connection-our-conven}
\end{table}

% learning, e.g., UDA~\cite{chen2021domain}, domain generalization (DG)~\cite{christiansen2021causal}, and out of distribution (OOD)~\cite{wang2022causal}.

\begin{figure*}[t]
	\setlength{\abovecaptionskip}{0pt}
	\begin{center}
		\includegraphics[width=0.95\linewidth]{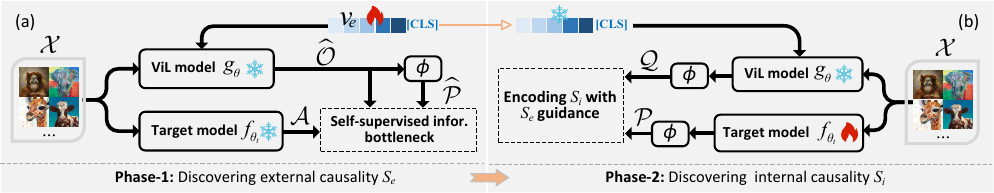} 
	\end{center}
	\caption{
        Overview of our {\modelnameshort} framework:
        (a) \texttt{Phase-1}: Discovering the external causal factors $S_e$ in form of prompt context $\boldsymbol{v}_{e}$
        from a frozen ViL model using our self-supervised information bottleneck algorithm;
        (b) \texttt{Phase-2}: Discovering the internal causal factors $S_i$
        where the updated prompt context $\boldsymbol{v}_{e}$ is used to predict pseudo-labels as the prior information.
        }
	\label{fig:fw}
\end{figure*}

An overview of our proposed {\modelnameshort} framework is illustrated in Fig.~\ref{fig:scm}(b). 
It models both non-causal factors $U$ and causal factors $S$  within a latent space ${A}$. 
Notably, the causal component $S$ is further decomposed into internal ($S_i$) and external ($S_e$) sub-components to facilitate a more complete and interpretable modeling of causality.

\changedrr{
This internal-external causality decomposition is motivated by the recognition that identifying all relevant causal relationships solely from the source and target domains is often infeasible due to limited training data and domain-specific knowledge. To address this, we introduce a pretrained ViL model, such as CLIP, as an external knowledge source to aid in the discovery of causal relationships.
This design leads to two types of causality: 
(i) {\bf Internal causality} referring to causal factors that can be inferred from the source domain model and the target domain data; (ii) {\bf External causality}, additional complements captured from the ViL model with broad, many-domain information. 
%that may not be fully observable within the dataset. In this paper, 
% \begin{itemize}
%     \setlength{\leftmargin}{0pt}
%     \item[(i)]  {\bf Internal causality} refers to causal factors that can be inferred directly from the observed data in the source and target domains. 
%     % These represent relationships that are learnable through conventional domain adaptation.
%     \item[(ii)] {\bf External causality}, in contrast, captures additional causality encoded in ViL models. These models provide broad, cross-domain information that can help fill in causal gaps not explicitly present in the training data.
% \end{itemize}
% }
% \changedrr{
%By modeling this dual-source causality, {\modelnameshort} leverages the complementary strengths of internal data-driven insights and external model-informed priors. 
Such a collaborative design enables a more holistic understanding of causal structures, as empirically validated in \texttt{Sec.\ref{sec:cau-vali}} and \texttt{Sec.\ref{sec:cau-ana}}. 
}

In contrast to conventional intervention-driven learning, we focus on discovering $S$ from the intertwined $S$ and $U$ as  
% , our casual SFDA can be 
\begin{equation}
    \label{eqn:intervention}
    \begin{split}
        P(\boldsymbol{y}|S) = P(\boldsymbol{y}|dis(S|\{U,S\})),~S=\{S_i, S_e\},
    \end{split}
\end{equation}
\changedrr{
where $dis(S|\{U,S\}))$ means conducting discover operation on $\{U,S\}$ to decompose $S$.
}
To make this process computationally tractable, we consider
\begin{principle}
\label{ppl-cmp}
\textit{{\bf \textit{Independent Causal Mechanisms (ICM) Principle~\cite{peters2017elements}}}: The conditional distribution of each variable given its causes (i.e., its mechanism) does not inform or influence the other mechanisms.}
\end{principle}

\changedxx{
As in \texttt{Sec.\ref{sec:causal-dg-uda}}, the lack of access to the source domain prevents the use of conventional causal learning methods.}
\changedrr{
Inspired by Principle~\ref{ppl-cmp}, we assume conditional independence between external and internal causal factors to enable a divide-and-conquer learning approach. Under this perspective, we propose the following factorized approximation as: 
}
% \changedrr{
% Inspired by Principle~\ref{ppl-cmp}, we assume conditional independence between external and internal causal factors to enable a divide-and-conquer learning approach. Under this perspective, we propose
% }
% Principle 1 offers a key insight: causal factors can be identified using a divide-and-conquer strategy due to their conditional independence. Thus, we propose a cumulative approach to discovering both external and internal causal factors iteratively:
\begin{equation}
    \label{eqn:prom-final}
    \small 
    \begin{split} 
        P(\boldsymbol{y}|S) \approx P(\boldsymbol{y}|dis(S_e|\{U,S_e\})) \cdot P(\boldsymbol{y}|dis(S_i|\{U,S_i\})). 
    \end{split}
\end{equation}

% \changedrr{
% \noindent{\bf Remark.}
% This factorization, as in Eq.~\eqref{eqn:prom-final}, is adopted as a {\it modeling assumption}, motivated by the spirit of the ICM principle, which states that, if such a decomposition is valid, the mechanisms do not inform each other. However, we emphasize that this factorization is not implied by the principle itself, but is introduced to enable tractable modeling and modular learning.
% }

% Based on the assumption that the causal factors $S_i$ and $S_e$ contribute independently to the prediction of $y$, we adopt a divide-and-conquer formulation as in Eq.~\eqref{eqn:prom-final}. This assumption aligns with the spirit of the ICM principle, which states that the mechanisms do not inform each other if such a decomposition is valid. 
% However, we emphasize that this factorization is not implied by the principle itself, but is introduced to enable tractable modeling and modular learning.

\subsection{{\modelnameshort} design}
We present a concrete design to carry out {\modelnameshort},
as depicted in Fig.~\ref{fig:fw}. 
By Eq.~\eqref{eqn:prom-final}, we implement the causality discovery in two successive phases: {\tt phase-1} for $S_e$
and {\tt phase-2} for $S_i$, as detailed below.

\begin{figure*}[t]
    \setlength{\abovecaptionskip}{0pt}
    \begin{center}
        \includegraphics[width=0.95\linewidth]{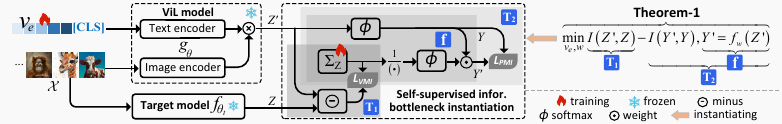} 
    \end{center}
    \caption{  
    \changedrr{Realizing} our self-supervised information bottleneck w.r.t. {\bf Theorem}~\ref{thm-one}. 
    $\rm{T}_1$ is estimated by \changed{Variational Mutual Information (VMI)} with Gaussian distribution assumption~$\mathcal{N}(Z, \Sigma_Z)$. 
    $\rm{T}_2$ is computed by \changed{Probabilistic Mutual Information (PMI)} where function $f_w$ integrates min-max optimization, using \changedrr{covariance} matrix $\Sigma_Z$ from $\rm{T}_1$ as weighting parameters $w$.     
    }
    \label{fig:SSIB-instan}
\end{figure*}

\subsubsection{External causality discovery} \label{sec:ccec-1}

To obtain external causality for a target task, we explore the potential of recent large multimodal ViL models such as CLIP~\cite{radford2021learning}.
\changedrr{
This is because they have accumulated a vast amount of knowledge and exhibited strong generalization abilities across a wide range of perception tasks. 
}
% This is because they have acquired a vast amount of knowledge in pertaining and exhibited general ability in many different perception tasks.

Specifically, we accomplish this idea by encoding $S_e$ into the prompt context $\boldsymbol{v}_e$ of a ViL model $g_\theta$ efficiently. 
That is, our learning target is $\boldsymbol{v}_e$, which is the explicit expression of $S_e$. 
We start by formulating the $S_e$ discovery as maximizing the correlation between a random variable representing $S_e$ and the prediction of the target model. 
Then, the original formulation is converted to a self-supervised information bottleneck problem with a theoretical guarantee, followed by the deep learning instantiation.

With the target samples $\mathcal{X}$ and learnable $\boldsymbol{v}_{e}$,
we obtain the prediction of  $g_\theta$ as $\hat{\mathcal{P}}=\{\hat{\boldsymbol{p}}_{i}\}_{i=1}^{n}$. 
Meanwhile, we obtain the logits of target model $f_{\theta_t}$ for the same data as ${\mathcal{A}}=\{ {\boldsymbol{a}}_i\}_{i=1}^{n}$.
We introduce three {\it random variables}, $V$, $Y$, and $Z$, following the probability distributions of $\boldsymbol{v}_{e}$, $\hat{\mathcal{P}}$ and ${\mathcal{A}}$, respectively. 
\changed{In {\modelnameshort}, we further write %$Z=\{Z_e, Z_i\}$ 
\begin{equation}
    Z=\{Z_e, Z_i\}
    \label{eqn:zzz}
\end{equation}
where the latent random variables $Z_{e}$ and $Z_{i}$ represent the external $S_{e}$ and the internal $S_{i}$ causal factors, respectively.
}

In an information-theoretic view, for causality relationships, we have to ensure the correlation between ${Z}_e$ and $V$. 
This could be considered an approximation with computational ease.
Formally, discovering $S_e$ needs maximizing the following mutual information:
\begin{equation}
	\label{eqn:prom-zdx}
	\begin{split}
		\max_{\boldsymbol{v}_e} I\left( {{Z_e},V} \right),
	\end{split}
\end{equation} 
where $I(\cdot, \cdot)$ is mutual information of two random variables.
\changedrr{Note, mutual information is not a prerequisite for causality discovery; instead, it is an effective design choice that enhances the discovery of external causality.}

% \changedrr{
% \noindent{\bf Remark.} The mutual information does not define causality in our framework, but when combined with the post-intervention assumption and external ViL prior, it serves as a practical signal for identifying potential causal factors. 
% }

\changed{
To achieve Eq.~\eqref{eqn:prom-zdx}, we employ prompt learning \cite{zhou2022coop} to accomplish the optimization. 
This is because it allows us to leverage CLIP without tuning model parameters. This avoids potential issues such as catastrophic forgetting and unintended distortion of the pretrained knowledge within the ViL model. 
Moreover, by preserving CLIP's pretrained knowledge, prompt learning facilitates the discovery of external causality, enabling more effective integration of external knowledge into the causal learning process. 
}

Consider that prompt $V$ is specific to the ViL model $g_{\theta}$, we further transform the above formula of Eq.~\eqref{eqn:prom-zdx} as
\begin{equation}
	\label{eqn:prom-zdy}
	\begin{split}
		\max_{\boldsymbol{v}_e} I\left( {{Z_e},Y} \right), Y=\phi \left( g_{\theta}\left( V \right) \right),
	\end{split}
\end{equation}
where $\phi(\cdot)$ is softmax function; so that the two variables under correlation maximization
reside in \changedrr{similar} semantic spaces.
\changedrr{Theoretically, we prove that this transformation can constitute a lower bound on the original, ensuring its validity in optimization}. 

First, we have the following {\bf Lemma}~\ref{lem-one}~(see the \changed{proof in \texttt{Appendix-A}}). 
\begin{lemma}
    \textit{Given random variables $Z_1$, $X_1$ and $Y_1$ where $X_1$, $Y_1$ satisfy a mapping $f_1: X_1 \mapsto Y_1$. When $f_1$ is compressed, i.e., the output's dimension is smaller than the input's, } 
    \begin{equation}
        \label{eqn:}
        \begin{split}
            I\left( {{Z_1},X_1} \right) \geq I\left( {{Z_1},Y_1} \right). 
        \end{split}
    \end{equation}
    \label{lem-one}
\end{lemma}

\vspace{-10pt}

The ViL model $g_{\theta}$ is indeed compressing, as it maps a high-dimensional image to a low-dimensional category vector. 
Let $Z_1=Z_e$, $X_1=X$, $Y_1=Y$ and $f_1 = \phi \left(g_{\theta}(\cdot) \right)$, we have $I({{Z_e},V}) \geq I( {{Z_e},Y})$ according to  {\bf Lemma}~\ref{lem-one}.

\vspace{2pt}
{\bf Self-supervised information bottleneck.}
In Eq.~\eqref{eqn:prom-zdy}, $Z_e$ is the hidden part of the observable \changedrr{logits} along with non-causal and internal causal factors. 
To facilitate this learning process of $Z_e$, we derive a self-supervised information bottleneck algorithm without prior assumption on the distribution~\cite{kawaguchi2023does}.
In general, information bottleneck conducts min-max optimization to squeeze out the essentials w.r.t. a task. 
{We introduce} a random variable $Z{'}$ (i.e., the bottleneck) for the ViL's logits $\widehat{\mathcal{O}}=\{ \hat{\boldsymbol{o}}_i\}_{i=1}^{n}$ between $V$ and $Y$, thereby forming a bottleneck style expression as:
\begin{equation}
    \label{eqn:ib-ori}
    \begin{split}
        \min_{\boldsymbol{v}_e} I\left( {Z,Z{'}} \right) - I\left( {Z{'},Y} \right),
    \end{split}
\end{equation}
where $Z$ is the logit variable of target model $f_{\theta_t}$ which is observed.
Note, we do not include the item $I\left( {Z{'},Y} \right)$ in minimization due to \changedrr{lacking supervision} signal like ground-truth labels as used in conventional information bottleneck~\cite{tishby2000information},
i.e., a self-supervised formulation. 
However, we theoretically prove that Eq.~\eqref{eqn:ib-ori} \changedrr{admits an upper bound, thus optimizationally valid}.

\begin{theorem}
% \begin{shaded}
\textit{Suppose that there are five random variables $Z$, $V$, $Z{'}$, $Y$ and $Y{'}$. Among them, $Z$ represents the target domain knowledge; $V$, $Z{'}$ and $Y$ express the input instances, an intermediate features of the ViL model and predictions, respectively; $Y{'}$ depicts a pseudo-label that has a functional relationship $f_w$~($w$ is learnable parameters) with $Z{'}$. When $f_w$ is reversible and uncompressed, we have this upper bound:} 
\begin{equation}
    \label{eqn:dsib}
    \small
    \begin{split}
    &\min_{\boldsymbol{v}_e} I\left( {Z,{Z{'}}} \right) - I\left( {{Z{'}},Y} \right) \\
    &\le \min_{\boldsymbol{v}_e,w} \underbrace {I\left( {Z,{Z{'}}} \right)}_{\rm{T_1}} - \underbrace {I\left( {{Y{'}},Y} \right),{Y{'}} = f_{w}\left( {{Z{'}}} \right)}_{\rm{T_2}}.
    \end{split}
\end{equation}
\label{thm-one}
% \end{shaded}
\end{theorem}

\vspace{-0.2cm}
The proof is \changed{given in \texttt{Appendix-B}}.
{\bf Theorem}~\ref{thm-one} suggests that the desired latent factors refinement is conditionally equivalent to a self-supervised information bottleneck, when pseudo-label $Y{'}$ is generated from the intermediate ViL feature $Z{'}$ without information loss.

We realize the proposed information bottleneck
using variational information maximization \cite{houthooft2016vime}.
The key is to construct the function $f_w$ with three properties:
(i) The output is a probability distribution, 
e.g., including a softmax operation $\phi$. 
(ii) No information loss, e.g., no zero weights in $w$ for \changedrr{weighting-based} functions.
(iii) The weighting parameters $w$ are constrained to be relevant to $I(Z{'},Z)$. 
The \changedrr{instantiations} of $T_1$ and $T_2$ are elaborated below.

As exactly computing mutual information of two vector variables is intractable, we maximize the variational lower bound for the term $T_1$~(see the proof in~\cite{barber2004algorithm}): 
\begin{equation}
    \label{eqn:dsib-t1}
    \small
    \begin{split}
    I\left( Z, Z{'}\right)
    & = \mathbb{E}_{{Z,{Z{'}}}} \left[ \log \frac{q\!\left( {Z{'}} | Z \right)}{p\left( {Z{'}} \right)} \right] + \displaystyle {\rm{KL}} \left( \left. {q\!\left( {Z{'}} | Z \right)} \right\| p\left( {Z{'}} | Z \right) \right) \\
    &\geq H\left( {Z{'}} \right) + \mathbb{E}_{{Z,{Z{'}}}} \left( \log q\left( {Z{'}} | Z \right) \right), 
    \end{split}
\end{equation} 
where $q( {Z{'}} | Z)$ is variational distribution approximating the real distribution $p({Z{'}} | Z )$, ${\rm{KL}}(\cdot,\cdot)$ is ${\rm{KL}}$-divergence function, and $H({Z{'}})$ is a constant. 
In practice, we model $q({Z{'}} | Z)$ as a Gaussian distribution 
$\mathcal{N}(Z, \Sigma_{Z})$ with mean $Z$ and diagonal covariance matrix $\Sigma_{Z}$. 
\changed{For minimizing $T_1$, we adopt the Variational Mutual Information (VMI) loss} (see $\rm{\mathbf{T}}_{1}$ in Fig.~\ref{fig:SSIB-instan}):
\begin{equation}
    \label{eqn:loss-vmi}
    \begin{split} 
        &\mathcal{L}_{\rm{VMI}}\left( \mathcal{X}; \boldsymbol{v}_e, \Sigma_{Z}\right)\\
        &=-\mathbb{E}_{{Z,{Z{'}}}} \left( ||{Z{'}} - Z ||_{\Sigma_{Z}^{-1}}^2 + \log |\Sigma_{Z}| \right)\\
        &= -\frac{1}{n}\sum_{i=1}^{n}\left( \hat{\boldsymbol{o}}_i - {\boldsymbol{a}}_{i}\right)^{T} {\Sigma_{Z}^{-1}}\left( \hat{\boldsymbol{o}}_i- {\boldsymbol{a}}_i\right) + \log{|\Sigma_{Z}^{i}|}.
    \end{split}
\end{equation}

Consider that the ViL's prediction $Y$ and the pseudo-label $Y{'}$ are probability distributions, we compute the $\rm{T_2}$ term by vanilla mutual information \cite{ji2019invariant}:
$I(Z, Z{'})={p(Z, Z{'})} \mathord{\left/
 {\vphantom {{} {}}} \right.
 \kern-\nulldelimiterspace} {[p(Z)p(Z{'})]}$.
\changed{The corresponding Probabilistic Mutual Information (PMI) loss} function is formed as:
\begin{equation}
    \label{eqn:loss-pmi}
    \small
    \begin{split}
    &\mathcal{L}_{\rm{PMI}}\left( \mathcal{X}; \boldsymbol{v}_e, \Sigma_{Z}\right)= \frac{1}{n}\sum_{i=1}^{n} {I}\left( \hat{\boldsymbol{p}}_{i}, {\boldsymbol{p}}'_{i}\right), \\
    &\text{with}\, \hat{\boldsymbol{p}}_{i}=\phi(\hat{\boldsymbol{o}}_i),~~{\boldsymbol{p}}'_{i}=f_w(\hat{\boldsymbol{o}}_i),
    \end{split}
\end{equation}
where \changedrr{$\sum_{c=1}^{C}{\boldsymbol{p}}'_{i,c}=1$}; $f_w(\cdot)$ is a weighting-based function constructed as follows.

From Eq.~\eqref{eqn:loss-vmi},
$I(Z{'},Z)$ is a Mean-Square Error (MSE) weighted by $\Sigma_Z$.
That means $\Sigma_Z$ detects dimension importance duration minimization, with high-weight dimensions to be suppressed, thus corresponding to the non-causal factors $U$ in our formulation.
Thus, taking ${{1} \mathord{\left/
 {\vphantom {{} {}}} \right.
 \kern-\nulldelimiterspace} {\Sigma_Z}}$ as the weighting parameters of $f_w$ shifts the focus of predicting
 to the causal component $S_e$ (see $\rm{\mathbf{f}}$ in Fig.~\ref{fig:SSIB-instan}), formally,
\changedrr{
\begin{equation}
    \label{eqn:f}
    \small
    \begin{split} 
        {\boldsymbol{p}}'_{i} = f_{\left[w=\Sigma_Z^{i}\right]}\left( \hat{\boldsymbol{o}}_{i}\right) = \phi \left( \phi \left(\frac{1}{\Sigma_Z^{i}}\right) \odot \phi\left( \hat{\boldsymbol{o}}_{i} \right) \right),
    \end{split}
\end{equation}
where $\phi(\cdot)$ is softmax function.} 
Combining Eq.~\eqref{eqn:loss-vmi},~\eqref{eqn:loss-pmi}, and~\eqref{eqn:f} together, our self-supervised objective for external causality discovery is formed as
\begin{equation}
    \label{eqn:loss-dsib}
    \small
    \begin{split} 
         % \left( \boldsymbol{\rm{SSIB}} \right)~&
         &\min_{{\boldsymbol{v}_e},\Sigma_{Z}} \mathcal{{L}}_{\rm{EC}}\left( \mathcal{X}; \boldsymbol{v}_e, \Sigma_{Z}\right)\\ 
         &= \mathcal{{L}}_{\rm{PMI}}\left( \mathcal{X}; \boldsymbol{v}_e, \Sigma_{Z}\right) - \alpha \mathcal{{L}}_{\rm{VMI}}\left( \mathcal{X}; \boldsymbol{v}_e, \Sigma_{Z}\right). 
    \end{split}
\end{equation}

\textbf{Remark}.
Our design differs from the existing information bottleneck in two aspects: While previous information extraction \cite{li2022invariant} is confined to a single model, ours is a cross-model design.  
Our formulation is self-supervised, in contrast to previous supervised designs.

\subsubsection{Internal causality discovery} \label{sec:ccec-2}

In {\tt phase-2}, we subsequently discover the internal causal factors $S_i$
by updating the target model.
That is, we encode $S_i$ in the target model, as it is latent in nature.
Similar to capturing the external factors $S_{e}$ (Eq.~(5)), this encoding can be formulated by maximizing the following conditional mutual information: 
\begin{equation}
    \label{eqn:zzts}
    \max_{\theta_t} I\left({{Z_i},Z_{\theta}} \right),~~{\rm{s.t.}}~\max_{\theta_t} I\left({{Z_e^*},Z_{\theta}} \right)
\end{equation} 
where random variable $Z_{\theta}$ represents target model parameters $\theta_t$, whilst $Z_i$ denotes random variable of internal causal factors $S_i$ (Eq.~(4)); 
$Z_e^*$ represents the currently discovered external causal factor $S_e^*$. \changedrr{The primary term, $\max_{\theta_t} I\left({{Z_i},Z_{\theta}} \right)$, aims to capture the internal causality $S_i$. The condition, $\max_{\theta_t} I\left({{Z_e^*},Z_{\theta}} \right)$, ensures the target model simultaneously accommodates the discovered external causality $S_e^*$.}

\changedrr{
Given ultra-high dimension and complexity of ${\theta_t}$, combined with the hidden nature of $Z_i$, directly optimizing or computing Eq.~\eqref{eqn:zzts} is intractable.
To address this, we interpret the interaction between
$Z_{\theta}$ and $Z_i$ as the inherent modeling process that transforms an input sample from pixel space into a predictive probability space. This perspective allows us to reformulate the problem as maximizing the mutual information between the target model's input and output, both of which are {\it observed} and {\it operational}.
Crucially, theoretical proofs demonstrate that mutual information between inputs and outputs is maximized when outputs are both highly confident and exhibit balanced class distributions \cite{krause2010discriminative}. Inspired by this theoretical finding, we maximize this originally non-computable mutual information by minimizing the following objective: 
\begin{equation}
    \label{eqn:loss-ssl-ins-lun}
    \small
    \begin{split}
    {\cal L}_{\rm{UN}}({\mathcal{P}}; \theta_t)&= -\sum_{i=1}^{n}\boldsymbol{p}_i\log\boldsymbol{p}_i + \tau \sum\limits_{c = 1}^C {{{{\rm{KL}}}}} \left( {\left. {{\varrho _c}} \right\|{1 \over C}} \right), \\
    \text{with } &\boldsymbol{p}_i \in {\mathcal{P}}=\phi(f_{\theta_t}(\mathcal{X})).
    \end{split}
\end{equation}
In this equation, ${\mathcal{P}}$ represents the predictions 
of $n$ training samples $\mathcal{X}$ generated by the target model $f_{\theta_t}$.
The first term minimizes the information entropy of each prediction $\boldsymbol{p}_i = [p_{i,c}]_{c=1}^C$.
The KL divergence term, on the other hand, regulates the predictive balance across all $C$ categories via pulling the empirical distribution 
${\varrho}_c = \frac{1}{n} \sum_{i=1}^{n}{p}_{i,c}$ towards a uniform distribution.
The trade-off parameter $\tau$ balances these two terms.
}

\changedrr{
Similarly, we convert the external causality $S_e^*$ from its prompt format $\boldsymbol{v}_e$ to pseudo-labels by evaluating the ViL model $g_{\theta}$:
${\mathcal{Q}}=\phi(g_{\theta}(\mathcal{X}, \boldsymbol{v}_e))$.
This conversion simplifies model optimization, as it enables the use of the conventional softmax cross-entropy loss, ${{\cal L}_{{\rm{SCE}}}}\left( {{\cal P},{\cal Q};{\theta _t}} \right)$. 
}

\changedrr{
Combining these elements, we optimize Eq.~\eqref{eqn:zzts} using the following objective function:
\begin{equation}
    \label{eqn:loss-ssl-ins}
    {\cal L}_{\rm{IC}} = \min_{\theta _t} {\cal L}_{\rm{UN}}({\cal P}; \theta_t) + \sigma {{\cal L}_{{\rm{SCE}}}}\left( {{\cal P},{\cal Q};{\theta _t}} \right)
\end{equation}
where $\sigma$ serves as another trade-off parameter.
}

\begin{algorithm}[t]
    % \caption{Pseudo-code of the {\modelnameshort} approach.}
    \caption{Training \modelnameshort}
    \label{alg:traing}
    \vspace{5pt}
    \textbf{Input}: Unlabelled target data $\mathcal{X}$, pre-trained source model $f_{\theta_s}$, a frozen ViL model $g_{\theta}$, max training iteration number $M$, prompt context $\boldsymbol{v}_e$. 
    % \#iteration of each epoch $I_e$
    \begin{algorithmic}[1]
        \STATE Initialize target model $f_{\theta_t}$(by $f_{\theta_s}$), $\boldsymbol{v}_e$(by fixed template)
        % \STATE Initialize $\boldsymbol{v}_e$ with a fixed template
        \FOR{iter $k=1$ to ${M}$}
        \STATE Sample a mini-batch $\mathcal{X}_{b}$ from $\mathcal{X}$.
        \STATE Discover external causal factors $S_e$: update $(\boldsymbol{v}_e, \Sigma_{Z})$ by optimizing $\mathcal{{L}}_{\rm{EC}}\left( \mathcal{X}_b; \boldsymbol{v}_e, \Sigma_{Z}\right)$, fixing $\mathcal{{L}}_{\rm{IC}}$.
        \STATE Convert updated $\boldsymbol{v}_e$ to target pseudo-labels for this batch $\mathcal{X}_{b}$, i.e., obtain ${\mathcal{Q}}_b=\phi(g_{\theta}(\mathcal{X}_b, \boldsymbol{v}_e)$.
        \STATE Discover internal causal factors $S_i$: update $f_{\theta_t}$ by optimizing $\mathcal{{L}}_{\rm{IC}}$, fixing $\mathcal{{L}}_{\rm{EC}}$.
        \ENDFOR
        \STATE \textbf{return:} The adapted model $f_{\theta_t}$.
    \end{algorithmic}
\end{algorithm}

\subsection{Model training} \label{sec:model-trainig}
We train {\modelnameshort} in an alternating manner.
In each iteration, %the training for 
we first optimize prompt context $\boldsymbol{v}_e$ with ${\cal L}_{{\rm{EC}}}$ whilst the target model is frozen (i.e., {\tt phase-1}),
followed by training the target model by ${\cal L}_{{\rm{IC}}}$ (i.e., {\tt phase-2}) while the prompt context is frozen.
The ViL model is always frozen throughout.
More details for training \changedrr{are} presented in Alg.~\ref{alg:traing}.

\section{Experiments}\label{sec:rlt}
\subsection{Implementation details} 

{\bf Source models.} 
Following~\cite{zhou2022coop}, for source-free out-of-distribution generalization (SF-OODG) we adopt the Pytorch built-in ResNet50 model (pre-trained on ImageNet) as the source model, taking ImageNet variations as evaluation datasets. 
For other settings, the source model is trained on the source domain in a supervised manner, same as~\cite{liang2020we,tang2022sclm}.

{\bf Networks.} 
Our {\modelnameshort} involves two parts: the ViL model and the target model. 
We choose CLIP~\cite{radford2021learning} as the ViL model. 
The target model's structure is the same as the source model.  
Following~\cite{liang2020we,yang2023trust}, 
a target model consists of a feature extractor and a classifier (an FC layer).
We adapt the feature extractor, which \changedrr{is} pre-trained on ImageNet. 
For fair comparison, we use ResNet101 as the backbone for VisDA and ResNet50 for the others.

{\bf Training.}
There are three parameters with {\modelnameshort}: $\alpha$ in $L_{\rm{EC}}$ (Eq.~\eqref{eqn:loss-dsib}), $\tau$ in $L_{\rm{UN}}$ (Eq.~\eqref{eqn:loss-ssl-ins-lun}), and $\sigma$ in $L_{\rm{IC}}$ (Eq.~\eqref{eqn:loss-ssl-ins}). 
For all settings, we adopt the same configuration $(\alpha, \sigma, \tau)=(0.003, 0.4, 1.0)$. Here, since the value of $\mathcal{{L}}_{\rm{VMI}}$ is large, $\alpha$ is set as small as 0.003.     
% {\bf Training setting.}
For model training, we use \changedrr{a batch} size of 64, \changedrr{an} SGD optimizer with momentum 0.9, and 15 epochs on all datasets. 
\changedxx{
The learnable prompt template is initialized with 'a photo of a [CLS].'~\cite{radford2021learning}, where the [CLS] means a specific target class name. 
}

\changedrrr{
{\bf Unification evaluation protocol}. 
In the SFDA context, unification does not necessarily imply that the proposed method must achieve the absolute best performance in every individual adaptation scenario. 
Instead, the essence of contributing a unified solution lies in its ability to provide stable and competitive performance across a wide range of settings in a unified manner, without the need of resorting to a diversity of scenario specific designs. 
}

\changedrrr{
To evaluate the unification of methods, we adopt three harmonic metrics: (1) Overall mean $H_{all}$ that averages over all settings, (2) worst-case relative gap $H_{wrg}$ that is largest percentage shortfall of a method from the per-setting best across all settings, and (3) leave-one-setting-out averages $H_{loso}$ omits one setting at a time and averages the scores over the rest.
For a specific method $m$ evaluated over a number of settings $\mathcal{S}$, the metrics can be computed by:
\begin{equation}
\label{eqn:unification-metrics}
\small
\begin{split}
&\mathrm{H}_{all}(m) = \frac{1}{|\mathcal{S}|}\sum_{s\in\mathcal{S}}x_{m,s};~
\mathrm{H}_{wrg}(m)
= \max_{s\in\mathcal{S}} \frac{b_s - x_{m,s}}{b_s};\\
&\mathrm{H}_{loso}(m, s') = \frac{1}{|\mathcal{S}-1|}\sum_{s\in\mathcal{S} \setminus s'}x_{m,s},
\end{split}
\end{equation}
where $x_{m,s}$ specifies the score of method $m$ under a setting $s$; $b_s\!\!=\!\!\max_{m'} x_{m',s}$ is the per-setting best; $s'$ is a omitted setting. 
}

\vspace{-0.2cm}
\subsection{\changedrrr{Indirect validation of causal factors}} \label{sec:cau-vali}

\changedrr{
Validating the discovery of causal factors is significant yet non-trivial in causality learning,
especially with latent causal factors as in this study.
%Directly examining the causal factors is an undoable task, since causality is a latent/unobserved relationship. 
To that end, we adopt a general principle that there exists an intimate connection between invariance and causality useful for generalization \cite{scholkopf2021toward,arjovsky2019invariant}.
That being said, if a learned model enables varying environments generalization, \changedrrr{it constitutes {\it indirect evidence} that the model has captured causal predictors (e.g., a causal explanation of an object, such as why it is a computer).}
}
% That being said, if a learned model enables varying environments generalization, it provides indirect evidence for that that causal predictors have been captured (e.g., causal explanation
% of an object such as why it is a computer).
%A truly invariant predictor—one that captures the causal mechanism
% In this paper, we propose to validate the causality by checking models' invariance, ground on the conclusion: A truly invariant predictor—one that captures the causal mechanism—should generalize well across different environments~\cite{scholkopf2021toward,arjovsky2019invariant}.

\begin{figure}[t]
    \begin{center}
        \includegraphics[width=0.4 \linewidth, height=0.38 \linewidth]{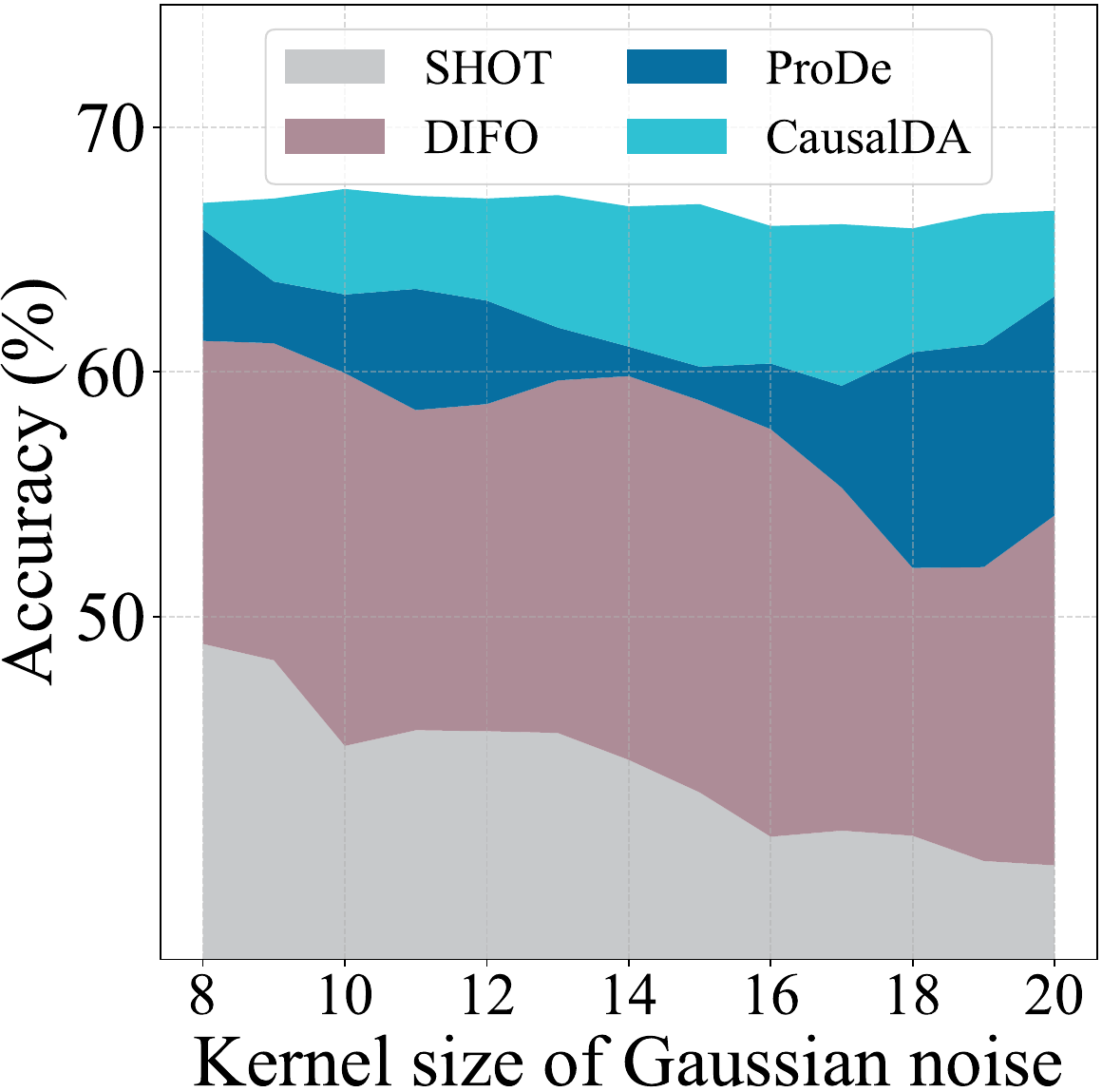}~~
        \includegraphics[width=0.4 \linewidth, height=0.38 \linewidth]{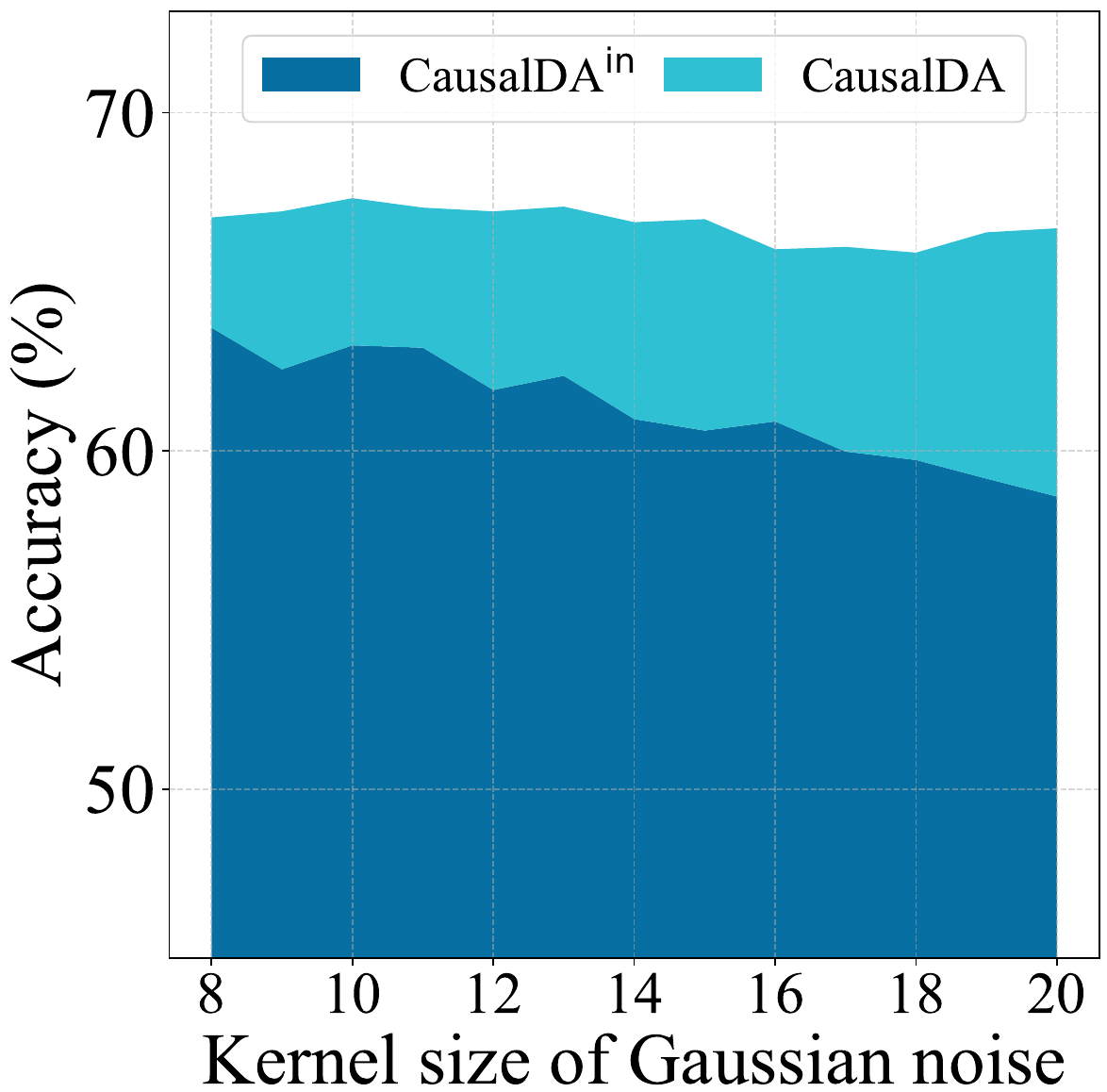}
    \end{center}
    \setlength{\abovecaptionskip}{0cm}
    \caption{
    \changedrr{Causality validation: Invariance analysis across varying noise settings on the Ar$\rightarrow$Cl task of {\bf Office-home}.}
    }
    \label{fig:cmp-noise}
\end{figure}

\changedrr{
Concretely, we simulate a spectrum of viewing environments
by imposing varying levels of Gaussian noise with gradually increased kernel sizes from 8 to 20.
%To this end, we perform a robustness analysis by applying Gaussian blur to the target images, gradually increasing the kernel size from 8 to 20, and measuring accuracy changes. 
%In these experiments, each kernel size is treated as a different environment. 
% {\modelnameshort} adopts ViT-B/32 [60] as the backbone 
We compare the competitors without (SHOT \cite{liang2020we}) and with (DIFO~\cite{tang2024source} and ProDe~\cite{tang2025proxy}) ViL knowledge \changedrrr{(the image-encoder in CLIP adopts backbone ViT-B/32~\cite{han2022survey})}.
As shown in Fig.~\ref{fig:cmp-noise}, all competitors degrade clearly as the noise increases, whilst {\modelnameshort} levels off.
\changedrrr{These findings suggest the method captures invariant predictors that are compatible with causal explanations, though they do not prove causal recovery rigorously.}
% This validates that our method has uniquely captured the underlying invariant predictors to make our model causal.
To further examine if external and internal factors both are effective, we further derive a variant of {\modelnameshort}, named as {\modelnameshort}$^{in}$, without the external causality discovery.
Specifically, in {\modelnameshort}$^{in}$, the pseudo-labels are directly derived from CLIP’s zero-shot predictions. 
% It is evident that external factors are useful in improving the invariance ability, confirming its complementarity with the internal counterparts.
\changedrrr{This is consistent with external factors contributing to improved invariance, suggesting a possible complementarity with internal counterparts.}
}

\changedrrr{The validation above provides indirect yet preliminary evidence for causality capture. In the following, we present further supporting evidence through adaptation results across diverse SFDA settings.}

\begin{table*}[t]
    \caption{
    Closed-set results (\%) on {\bf Office-Home}, {\bf VisDA}. 
    \changed{{SF} and {ViL} means source-free and ViL model-required, respectively.}
    }
    \label{tab:oh-vc-sfda}
    \renewcommand\tabcolsep{1.5pt}
    \renewcommand\arraystretch{0.9} 
    % \footnotesize
    % \small
    \scriptsize
    \centering
    \begin{tabular}{ l l c c c c c c c c c c c c c c c | c}
        \toprule
        \multirow{2}{*}{Method} &\multirow{2}{*}{Venue} &\multirow{2}{*}{SF}
        &\multirow{2}{*}{ViL}
        &\multicolumn{13}{c}{\textbf{Office-Home}} \vline &{\textbf{VisDA}} \\
        & & &&Ar$\to$Cl &Ar$\to$Pr &Ar$\to$Rw
            &Cl$\to$Ar &Cl$\to$Pr &Cl$\to$Rw    
            &Pr$\to$Ar &Pr$\to$Cl &Pr$\to$Rw  
            &Rw$\to$Ar &Rw$\to$Cl &Rw$\to$Pr &Avg. &Sy$\to$Re.\\
        \midrule
        Source    &--      &-- &--           &43.7 &67.0 &73.9 &49.9 &60.1 &62.5 &51.7 &40.9 &72.6 &64.2 &46.3 &78.1 &59.2 &49.2 \\
        CLIP-RN~\cite{radford2021learning}   &ICML21  &-- &\cmark   &51.9 &81.5 &82.5 &72.5 &81.5 &82.5 &72.5 &51.9 &82.5 &72.5 &51.9 &81.5 &72.1 &83.7 \\
        % CLIP-B32~\cite{radford2021learning}  &ICML21  &-- &\cmark   &00.0 &00.0 &00.0 &00.0 &00.0 &00.0 &00.0 &00.0 &00.0 &00.0 &00.0 &00.0 &00.0 &00.0 \\
        % CLIP-B32~\cite{radford2021learning}  &ICML21  &-- &\cmark   &59.8 &84.3 &85.5 &74.6 &84.3 &85.5 &74.6 &59.8 &85.5 &74.6 &59.8 &84.3 &76.1 &82.9 \\

        CLIP-B32~\cite{radford2021learning}  &ICML21  &-- &\cmark &64.9	&86.6	&87.2	&78.0	&86.6 &87.2	&78.0	&64.9	&87.2	&78.0	&64.9	&86.6	&79.2	&85.2 \\
        \midrule
        DAPL-RN~\cite{ge2022domain}       &TNNLS23 &\xmark &\cmark    &54.1 &84.3 &84.8 &74.4 &83.7 &85.0 &74.5 &54.6 &84.8 &75.2 &54.7 &83.8 &74.5 &86.9 \\
        PADCLIP-RN~\cite{lai2023padclip}  &ICCV23  &\xmark  &\cmark   &57.5 &84.0 &83.8 &77.8 &85.5 &84.7 &76.3 &59.2 &85.4 &78.1 &60.2 &86.7 &76.6 &88.5 \\
        ADCLIP-RN~\cite{singha2023ad}     &ICCV23W  &\xmark  &\cmark  &55.4 &85.2 &85.6 &76.1 &85.8 &86.2 &76.7 &56.1 &85.4 &76.8 &56.1 &85.5 &75.9 &87.7 \\
        \midrule
        SHOT~\cite{liang2020we}         &ICML20  &\cmark   &\xmark  &55.0 &78.7 &81.3 &69.1 &78.9 &79.1 &68.2 &53.6 &81.6 &73.5 &59.4 &83.5 &71.8 &82.9 \\
        GKD~\cite{tang2021model}        &IROS21  &\cmark   &\xmark  &56.5 &78.3 &82.2 &69.2 &80.4 &78.7 &67.4 &55.4 &82.6 &74.3 &60.3 &84.2 &72.5 &83.0 \\
        NRC~\cite{yang2021nrc}          &NeurIPS21  &\cmark   &\xmark  &57.2 &79.3 &81.3 &68.9 &80.6 &80.2 &66.6 &57.3 &82.0 &71.0 &57.9 &84.9 &72.3 &85.9 \\
        AaD~\cite{yang2022aad}          &NeurIPS22  &\cmark   &\xmark  &59.3 &79.3 &82.1 &68.9 &79.8 &79.5 &67.2 &57.4 &83.1 &72.1 &58.5 &85.4 &72.7 &88.0 \\
        AdaCon~\cite{chen2022contrastive}     &CVPR22 &\cmark&\xmark &47.2 &75.1 &75.5 &60.7 &73.3 &73.2 &60.2 &45.2 &76.6 &65.6 &48.3 &79.1 &65.0 &86.8 \\
        CoWA~\cite{lee2022confidence}         &ICML22 &\cmark&\xmark &57.3 &79.3 &81.0 &69.3 &77.9 &79.6 &68.1 &56.4 &82.6 &72.9 &61.3 &83.7 &72.4 &86.9 \\ 
        PLUE~\cite{litrico2023guiding}        &CVPR23 &\cmark &\xmark &49.1 &73.5 &78.2 &62.9 &73.5 &74.5 &62.2 &48.3 &78.6 &68.6 &51.8 &81.5 &66.9 &88.3 \\
        TPDS~\cite{tang2023tpds}              &IJCV24 &\cmark &\xmark &59.3 &80.3 &82.1 &70.6 &79.4 &80.9 &69.8 &56.8 &82.1 &74.5 &61.2 &85.3 &73.5 &87.6 \\
        \midrule
        SHOT-B32~\cite{liang2020we}    &ICML20     &\cmark &\cmark   &64.7	 &84.0	&86.4	&79.4	&87.3	&85.4	&78.1	&65.8	&87.6	&81.6	&65.0	&88.5   &79.5  &85.2 \\
        GKD-B32~\cite{tang2021model}   &IROS21     &\cmark &\cmark   &67.2	 &83.2	&84.3	&77.0	&84.8	&83.6	&75.5	&71.1	&86.0	&81.1	&72.4	&88.5	&79.6  &85.3 \\
        NRC-B32~\cite{yang2021nrc}     &NeurIPS21  &\cmark &\cmark   &55.0	 &83.4	&86.7	&78.3	&86.5	&85.1	&77.1	&69.9	&87.2	&\textbf{\color{cmblu}90.9}	&65.5	&88.4	&79.5  &88.2 \\
        AaD-B32~\cite{yang2022aad}     &NeurIPS22  &\cmark &\cmark   &69.4	 &83.0	&84.9	&76.4	&85.2	&84.6	&73.3	&71.1	&86.3	&79.0	&72.1	&88.7	&79.5  &82.6 \\
        TPDS-B32~\cite{tang2023tpds}   &IJCV24     &\cmark &\cmark   &67.7	 &82.7	&86.1	&76.5	&88.0	&86.3	&73.6	&69.4	&87.1	&79.3	&70.5	&89.7	&79.7  &78.0 \\
        MMGA-B32~\cite{chen2025source} &PR25        &\cmark &\cmark  &68.5 &84.6 &86.8 &78.8 &87.5 &86.5 &75.9 &68.3 &86.3 &80.4 &69.9 &88.6 &80.2 &89.2\\
        \rowcolor{gray! 40} {\bf {\modelnameshort}}-C-RN  &-- &\cmark&\cmark &60.1 &85.6 
        &86.2 &77.2 &86.0 &86.3 &76.6 &61.0 &86.5 &77.5 &61.4 &86.2 &77.6 &89.3 \\
        \rowcolor{gray!40} {\bf {\modelnameshort}}-C-B32 &-- &\cmark &\cmark &\textbf{\color{cmblu}74.2} &\textbf{\color{cmblu}90.8} &\textbf{\color{cmblu}90.3} &\textbf{\color{cmblu}82.3} &\textbf{\color{cmblu}91.0} &\textbf{\color{cmblu}89.9} &\textbf{\color{cmblu}81.0} &\textbf{\color{cmblu}74.8} &\textbf{\color{cmblu}90.0} &82.5 &\textbf{\color{cmblu}74.1} &\textbf{\color{cmblu}90.9} &\textbf{\color{cmblu}84.3} &\textbf{\color{cmblu}90.3}\\
        % \rowcolor{gray! 40} {\bf {\modelnameshort}}-C-B32  &-- &\cmark &\cmark &\textbf{\color{cmblu}72.3} &\textbf{\color{cmblu}89.8} &\textbf{\color{cmblu}89.9} &\textbf{\color{cmblu}81.1} &\textbf{\color{cmblu}90.3} &\textbf{\color{cmblu}89.5} &\textbf{\color{cmblu}80.1} &\textbf{\color{cmblu}71.5} &\textbf{\color{cmblu}89.8} &\textbf{\color{cmblu}81.8} &\textbf{\color{cmblu}72.7} &\textbf{\color{cmblu}90.4} &\textbf{\color{cmblu}83.3} &\textbf{\color{cmblu}89.3} \\
        \bottomrule
    \end{tabular}
\end{table*}

\begin{table*}[t]
    \caption{Closed-set results (\%) on {\bf DomainNet-126}. 
    \changed{{SF} and {ViL} means source-free and ViL model-required, respectively.}
    }
    \label{tab:126-sfda}
    \renewcommand\tabcolsep{5.2pt}
    \renewcommand\arraystretch{0.9} 
    % \footnotesize
    % \small
    \scriptsize
    \centering
    \begin{tabular}{ l l c c c c c c c c c c c c c c c }
        \toprule
        Method &{Venue} &{SF} &{ViL}
        &C$\to$P &C$\to$R &C$\to$S
        &P$\to$C &P$\to$R &P$\to$S    
        &R$\to$C &R$\to$P &R$\to$S  
        &S$\to$C &S$\to$P &S$\to$R &Avg.\\
        \midrule
        Source      &--     &-- &--     &44.6 &59.8 &47.5 &53.3 &75.3 &46.2 &55.3 &62.7 &46.4 &55.1 &50.7 &59.5 &54.7 \\
        CLIP-RN~\cite{radford2021learning}     &ICML21 &-- &\cmark &70.2 &87.1 &65.4 &67.9 &87.1 &65.4 &67.9 &70.2 &65.4 &67.9 &70.2 &87.1 &72.7 \\
        % CLIP-B32~\cite{radford2021learning}    &ICML21 &-- &\cmark &00.0 &00.0 &00.0 &00.0 &00.0 &00.0 &00.0 &00.0 &00.0 &00.0 &00.0 &00.0 &00.0 \\
        % CLIP-B32~\cite{radford2021learning}    &ICML21 &-- &\cmark &73.5 &85.7 &71.2 &74.7 &85.7 &71.2 &74.7 &73.5 &71.2 &74.7 &73.5 &85.7 &76.3 \\
        % CLIP-B32~\cite{radford2021learning}    &ICML21 &-- &\cmark &77.2	&90.0	&74.6	&78.9	&90	&74.6	&78.9	&77.2	&74.6	&78.9	&77.2	&90.0	&80.2\\
        CLIP-B32~\cite{radford2021learning}    &ICML21 &-- &\cmark &76.2	&89.0	&73.6	&77.9	&89.0	&73.6	&77.9	&76.2	&73.6	&77.9	&76.2	&89.0	&79.2\\    
        \midrule
        DAPL-RN~\cite{ge2022domain}   &TNNLS23  &\xmark &\cmark  &72.4 &87.6 &65.9 &72.7 &87.6 &65.6 &73.2 &72.4 &66.2 &73.8 &72.9 &87.8 &74.8 \\ 
        ADCLIP-RN~\cite{singha2023ad} &ICCV23W  &\xmark &\cmark  &71.7 &88.1 &66.0 &73.2 &86.9 &65.2 &73.6 &73.0 &68.4 &72.3 &74.2 &89.3 &75.2 \\ 
        \midrule
        SHOT~\cite{liang2020we}     &ICML20 &\cmark &\xmark &63.5 &78.2 &59.5 &67.9 &81.3 &61.7 &67.7 &67.6 &57.8 &70.2 &64.0 &78.0 &68.1 \\
        GKD~\cite{tang2021model}    &IROS21 &\cmark &\xmark &61.4 &77.4 &60.3 &69.6 &81.4 &63.2 &68.3 &68.4 &59.5 &71.5 &65.2 &77.6 &68.7 \\ 
        NRC~\cite{yang2021nrc}      &NeurIPS21 &\cmark &\xmark &62.6 &77.1 &58.3 &62.9 &81.3 &60.7 &64.7 &69.4 &58.7 &69.4 &65.8 &78.7 &67.5 \\
        AdaCon~\cite{chen2022contrastive}   &CVPR22 &\cmark &\xmark &60.8 &74.8 &55.9 &62.2 &78.3 &58.2 &63.1 &68.1 &55.6 &67.1 &66.0 &75.4 &65.4 \\
        CoWA~\cite{lee2022confidence}       &ICML22 &\cmark &\xmark &64.6 &80.6 &60.6 &66.2 &79.8 &60.8 &69.0 &67.2 &60.0 &69.0 &65.8 &79.9 &68.6\\
        PLUE~\cite{litrico2023guiding}      &CVPR23 &\cmark &\xmark &59.8 &74.0 &56.0 &61.6 &78.5 &57.9 &61.6 &65.9 &53.8 &67.5 &64.3 &76.0 &64.7  \\
        TPDS~\cite{tang2023tpds}            &IJCV24 &\cmark &\xmark &62.9 &77.1 &59.8 &65.6 &79.0 &61.5 &66.4 &67.0 &58.2 &68.6 &64.3 &75.3 &67.1\\
        \midrule
        SHOT-B32~\cite{liang2020we}   &ICML20      &\cmark   &\cmark &72.8   &84.5 	&72.0 	&75.1 	&85.2 	&72.2 	&78.2 	&75.5 	&71.6 	&75.2 	&73.6 	&83.2 	&76.6 \\
        GKD-B32~\cite{tang2021model}  &IROS21      &\cmark   &\cmark &73.7   &84.4 	&73.6 	&76.1 	&85.4 	&73.6 	&80.4 	&76.9 	&73.9 	&74.9 	&74.8 	&82.1 	&77.5 \\
        NRC-B32~\cite{yang2021nrc}    &NeurIPS21   &\cmark   &\cmark &72.2   &83.4 	&72.5 	&75.1 	&84.8 	&72.6 	&78.3 	&76.7 	&74.0 	&75.0 	&73.7 	&82.6 	&76.7 \\
        AaD-B32~\cite{yang2022aad}    &NeurIPS22   &\cmark   &\cmark &72.2   &84.2 	&72.0 	&75.3 	&85.3 	&72.6 	&78.2 	&77.4 	&73.6 	&74.8 	&74.7 	&82.8 	&76.9 \\
        TPDS-B32~\cite{tang2023tpds}  &IJCV24      &\cmark   &\cmark &72.5   &82.8 	&71.8 	&77.7 	&84.7 	&72.9 	&\textbf{\color{cmblu}81.9} &76.5 	&72.8 	&73.5 	&74.2 	&83.0 	&77.0 \\
        MMGA-B32~\cite{chen2025source}    &PR25       &\cmark &\cmark  &75.4 &86.5 &69.9 &74.6 &88.3 &72.3 &75.9 &76.1 &70.9 &76.2 &75.6 &87.4 &77.4 \\
        \rowcolor{gray! 40}  {\bf {\modelnameshort}}-C-RN &--&\cmark &\cmark 
        &75.4 &88.2 &72.0 &75.8 &88.3 &72.1 &76.1 &75.6 &71.2 &77.6 &75.9 &88.2 &78.0 \\

        \rowcolor{gray!40} {\bf {\modelnameshort}}-C-B32 &--
        &\cmark &\cmark
        &\textbf{\color{cmblu}79.5} &\textbf{\color{cmblu}89.8} &\textbf{\color{cmblu}76.9} &\textbf{\color{cmblu}81.6} &\textbf{\color{cmblu}90.1} &\textbf{\color{cmblu}77.5} &81.7 &\textbf{\color{cmblu}80.4} &\textbf{\color{cmblu}76.6} &\textbf{\color{cmblu}82.5} &\textbf{\color{cmblu}80.1} &\textbf{\color{cmblu}89.9} &\textbf{\color{cmblu}82.2} \\
        
        % \rowcolor{gray! 40}  {\bf {\modelnameshort}}-C-B32 &--
        % &\cmark &\cmark
        % &\textbf{\color{cmblu}77.2} &88.0 &\textbf{\color{cmblu}75.2} &\textbf{\color{cmblu}78.8} &88.2 &\textbf{\color{cmblu}75.8} &\textbf{\color{cmblu}79.1} &\textbf{\color{cmblu}77.8} &\textbf{\color{cmblu}74.9} &\textbf{\color{cmblu}79.9} &\textbf{\color{cmblu}77.4} &88.0&\textbf{\color{cmblu}80.0} \\

        \bottomrule
    \end{tabular}
\end{table*}

\subsection{Closed-set SFDA} \label{sec:rlt-close-set}
\textbf{Datasets}. The evaluation is based on three challenging benchmarks.
\textbf{Office-Home}~\cite{venkateswara2017deep} is a middle-scale dataset. It contains 15k images belonging to 65 categories from working or family environments, being divided into four domains, i.e., {\bf Ar}tistic, {\bf Cl}ip Art, {\bf Pr}oduct, and {\bf R}eal-{\bf w}ord.
\textbf{VisDA}~\cite{peng2017visda} is a large-scale dataset. It includes 12 types of {\bf Sy}nthetic to {\bf Re}al transfer recognition tasks. The source domain contains 152k synthetic images, whilst the target domain has 55k real object images from COCO.
\textbf{DomainNet-126}~\cite{peng2019moment} is another large-scale dataset.  
As a subset of DomainNet containing 600k images of 345 classes from 6 domains of different image styles, this dataset has 145k images from 126 classes, sampled from 4 domains, {\bf C}lipart, {\bf P}ainting, {\bf R}eal, {\bf S}ketch, as \cite{peng2019moment} identifies severe noisy labels in the dataset.  
\begin{table}[t]
	\caption{One step adapting results (\%) on \textbf{Office-Home} in Generalized SFDA setting. {S}, {T}, {H} are the results on the source and target domains, and harmonic mean accuracy, respectively; the bracket values in red are the gap between {S} and {T}; WAD means With Anti-forgetting Design.}
	\label{tab:oh-st}
	\renewcommand\tabcolsep{8.5pt}
	\renewcommand\arraystretch{0.9}
	\scriptsize
	\centering
 	\begin{tabular}{ l l c c c c}
        \toprule
        \multirow{2}{*}{Method} &\multirow{2}{*}{Venue}  &\multirow{2}{*}{WAD} &\multicolumn{3}{c}{Avg.}\\
        \cline{4-6}
        & & &S &T &H \\
        \midrule
        Source &--  &\xmark &98.1 	&59.2	&73.1 \\
        \midrule
	    SHOT~\cite{liang2020we}            &ICML20    &\xmark &84.2  &71.8 &77.1 \\
	    GKD~\cite{tang2021model}           &IROS21    &\xmark &86.8  &72.5 &78.7 \\
        NRC~\cite{yang2021nrc}             &NeurIPS21 &\xmark &91.3  &72.3 &80.4 \\
        AdaCon~\cite{chen2022contrastive}  &CVPR22    &\xmark &88.2  &65.0 &74.4 \\
        CoWA~\cite{lee2022confidence}      &ICML22    &\xmark &91.8  &72.4 &80.6 \\
        PLUE~\cite{litrico2023guiding}     &CVPR23    &\xmark &96.3 &66.9 &78.4 \\
        TPDS~\cite{tang2023tpds}           &IJCV24    &\xmark &83.8  &73.5 &78.0 \\
        % \midrule
        GDA~\cite{yang2021generalized}     &ICCV21    &\cmark &80.0  &70.2  &74.4 \\
        PSAT-RN~\cite{tang2023psat}        &TMM24     &\cmark &85.3  &72.6  &78.4 \\
        \midrule
        SHOT-B32~\cite{liang2020we}      &ICML20      &\xmark &81.9 &79.5  &80.7 \\
        GKD-B32~\cite{tang2021model}     &IROS21      &\xmark &89.2 &79.5  &84.1 \\
        NRC-B32~\cite{yang2021nrc}       &NeurIPS21   &\xmark &81.1 &78.6  &79.8 \\
        TPDS-B32~\cite{tang2023tpds}     &IJCV24      &\xmark &88.9 &79.7  &84.1 \\
        PSAT-B32~\cite{tang2023psat}     &TMM24       &\cmark &89.5 &80.7  &84.9 \\
        % DIFO-B32~\cite{tang2024source}   &CVPR2024  &\xmark &78.0 &83.1 ({\color{cmred}5.1}) &80.5 \\
        % ProDe-B32~\cite{tang2024proxy}   &ICLR2025  &\cmark &84.1 &84.5 ({\color{cmred}0.4}) &84.3 \\
        MMGA-B32~\cite{chen2025source}     &PR25         &\xmark &86.9 &80.2  &83.4 \\
        \rowcolor{gray! 40} {\bf {\modelnameshort}}-C-RN   &--  &\xmark &85.0 &77.6  &80.7 \\ 
        \rowcolor{gray! 40} {\bf {\modelnameshort}}-C-B32  &--  &\xmark &86.1 &84.3  &\textbf{\color{cmblu}85.2} \\
        % \rowcolor{gray! 40} {\bf {\modelnameshort}}-C-B32  &--  &\xmark &86.3 &\textbf{\color{cmblu}83.3}  &84.5 \\
        % \rowcolor{gray! 40} {\bf {\modelnameshort}}-C-B32  &--  &\xmark &86.3 &\textbf{\color{cmblu}83.3}  &84.5 \\
        % &86.1 &84.3  &\textbf{\color{cmblu}85.2}
        % \midrule
        % \rowcolor{gray! 40} {\bf {\modelnameshort}}-C-B16  &--  &\xmark &86.3 &\textbf{\color{cmblu}85.2} ({\color{cmred}1.1})  &\textbf{\color{cmblu}85.8} \\
        \bottomrule
  	\end{tabular}
\end{table}

\begin{table*}[t]
    \caption{Continual adaptation results (\%) on \textbf{Office-Home} under Generalized SFDA setting, the first column of each sub-table indicates the adaptation sequence. $\downarrow$ means the average accuracy drop of a test domain on the adaptation path compared with the performance when the domain is first seen.}
	\label{tab:cda-contin}
	\renewcommand\tabcolsep{5.5pt}
	\renewcommand\arraystretch{0.9}
	% \footnotesize
        \scriptsize
	\centering
	\begin{tabular}{ |c|c|c|c|c|c|c|c|c|c|c|c|c|c|c|c|c|c|c|c|c|c|c|c|c c c c c c}
            \hline
            \multicolumn{23}{|c|}{{\modelnameshort}-C-RN} \\
            \hline
		\cline{1-5}
		\cline{7-11}
		\cline{13-17}
		\cline{19-23}
		\multirow{2}*{} &\multicolumn{4}{c|}{Test} & &\multirow{2}*{} &\multicolumn{4}{c|}{Test} & &\multirow{2}*{} &\multicolumn{4}{c|}{Test} & &\multirow{2}*{} &\multicolumn{4}{c|}{Test} \\
		\cline{2-5}
		\cline{8-11}
		\cline{14-17}
		\cline{20-23}
		&Ar &Cl &Pr &Rw & & &Cl &Ar &Pr &Rw & & &Pr &Ar &Cl &Rw & & &Rw &Ar &Cl &Pr \\
		\cline{1-5}
		\cline{7-11}
		\cline{13-17}
		\cline{19-23}
		    Ar    &97.8	 &46.1	&68.4  &71.1
            &  &Cl    &97.5	 &52.5	&65.8  &61.5
            &  &Pr    &99.6	 &52.2	&40.8  &71.3
            &  &Rw    &98.1	 &63.1	&46.3  &78.5 \\
            
		\cline{1-5}
		\cline{7-11}
		\cline{13-17}
		\cline{19-23}
		    Cl    &80.0 &63.5 &64.3 &66.0 
            &  &Ar    &80.1 &77.8 &70.7 &75.6 
            &  &Ar	  &91.2 &78.4 &50.2 &78.3 
            &  &Ar	  &93.4 &78.4 &50.4 &75.6  \\
		\cline{1-5}
		\cline{7-11}
		\cline{13-17}
		\cline{19-23}
		    Pr	 &78.4 &61.7 &85.0 &76.2 
    	&	&Pr	  &75.4 &74.4 &86.9 &78.1 

    	&	&Cl	  &81.5 &71.9 &61.5 &69.9 

    	&   &Cl	  &83.0 &74.7 &62.1 &69.7 \\
		\cline{1-5}
		\cline{7-11}
		\cline{13-17}
		\cline{19-23}
		      Rw	&83.1 &60.2 &83.2 &84.4 
	    &   &Rw   &67.6 &78.4 &85.4 &85.6 
            &   &Rw   &87.3 &76.9 &59.2 &85.6 
            &	&Pr	  &87.3 &72.5 &61.5 &85.2 \\
		\cline{1-5}
		\cline{7-11}
		\cline{13-17}
		\cline{19-23}
            $\downarrow$	   &17.3 &2.5 &1.8 &--
	    &   &$\downarrow$  &23.1 &1.4 &1.6 &--
            &   &$\downarrow$  &13.0 &4.1 &2.3 &--
            &	&$\downarrow$  &10.2 &4.8 &0.6 &-- \\
            \cline{1-5}
		\cline{7-11}
		\cline{13-17}
		\cline{19-23}
            \hline
            \end{tabular}
            \begin{tabular}{ |c|c|c|c|c|c|c|c|c|c|c|c|c|c|c|c|c|c|c|c|c|c|c|c|c c c c c c}
            \hline
            \multicolumn{23}{|c|}{{\modelnameshort}-C-B32} \\
            \hline
		\cline{1-5}
		\cline{7-11}
		\cline{13-17}
		\cline{19-23}
		\multirow{2}*{} &\multicolumn{4}{c|}{Test} & &\multirow{2}*{} &\multicolumn{4}{c|}{Test} & &\multirow{2}*{} &\multicolumn{4}{c|}{Test} & &\multirow{2}*{} &\multicolumn{4}{c|}{Test} \\
		\cline{2-5}
		\cline{8-11}
		\cline{14-17}
		\cline{20-23}
		&Ar &Cl &Pr &Rw & & &Cl &Ar &Pr &Rw & & &Pr &Ar &Cl &Rw & & &Rw &Ar &Cl &Pr \\
		\cline{1-5}
		\cline{7-11}
		\cline{13-17}
		\cline{19-23}
		    Ar    &97.8	 &46.1	&68.4  &71.1
            &  &Cl    &97.5	 &52.5	&65.8  &61.5
            &  &Pr    &99.6	 &52.2	&40.8  &71.3
            &  &Rw    &98.1	 &63.1	&46.3  &78.5 \\
            
		\cline{1-5}
		\cline{7-11}
		\cline{13-17}
		\cline{19-23}
		    Cl    &84.1	&77.0 &69.3 &72.5
            &  &Ar    &80.9	&83.4 &72.3 &77.0
            &  &Ar	  &90.0 &82.5 &49.4	&80.1
            &  &Ar	  &94.0	&82.8 &51.2	&76.8 \\
		\cline{1-5}
		\cline{7-11}
		\cline{13-17}
		\cline{19-23}
		    Pr	    &82.5   &70.9   &91.0	&81.1
    	&	&Pr	    &74.2	&78.1	&90.4	&82.2
    	&	&Cl	    &83.6	&75.6	&76.2	&76.2
    	&   &Cl	    &85.2	&73.8	&76.2	&72.5 \\
		\cline{1-5}
		\cline{7-11}
		\cline{13-17}
		\cline{19-23}
		      Rw	&86.3	&67.6	&88.3	&87.7
	    &   &Rw   &69.5	  &81.3	  &88.9	  &89.1
            &   &Rw   &89.8	  &79.1	  &69.5	  &87.7
            &	&Pr	  &89.5	  &76.9	  &71.5	  &90.0 \\
		\cline{1-5}
		\cline{7-11}
		\cline{13-17}
		\cline{19-23}
            $\downarrow$	   &13.5  &7.7	 &2.7	&--
	    &   &$\downarrow$  &22.6  &3.7	 &1.5	&--
            &   &$\downarrow$  &11.8  &5.2	 &6.7	&--
            &	&$\downarrow$  &8.5	  &7.5	 &4.7	&-- \\
            \cline{1-5}
		\cline{7-11}
		\cline{13-17}
		\cline{19-23}
            \hline
	\end{tabular}
\end{table*}

\textbf{Competitors}. 
To evaluate {\modelnameshort}, we select 14 state-of-the-art methods in three groups. 
{\itshape (i) The first} is the source model and CLIP that lay the basis for {\modelnameshort}. 
% Among them, method ResNet-50 and ResNet-101 are used to initiate the feature extractor of the source model, whilst No-adaptation's results are yielded by the source model only. 
{\itshape (ii) The second} is the UDA method 
DAPL~\cite{ge2022domain}, 
PADCLIP~\cite{lai2023padclip}, 
ADCLIP~\cite{singha2023ad} 
that introduces prompt learning to boost the cross-domain transfer. 
{\itshape (iii) The third} includes 9 SFDA methods:
SHOT~\cite{liang2020we},
GKD~\cite{tang2021model},
NRC~\cite{yang2021nrc}, 
AaD~\cite{yang2022aad}, 
% SFDE~\cite{ding2022source}, 
AdaCon~\cite{chen2022contrastive},
CoWA~\cite{lee2022confidence}, 
% ELR~\cite{yi2023source}, 
PLUE~\cite{litrico2023guiding},
TPDS~\cite{tang2023tpds},
and 
MMGA~\cite{chen2025source}.
By idea, SHOT, GKD, CoWA are pseudo-labels-based methods; NRC, AaD, PLUE, AdaCon exploit the data geometry, e.g., the nearest neighbor, neighborhood, prototypes; TPDS predicts the target probability distribution using error control; MMGA aligns the target domain with the multimodal space represented by CLIP.

For comprehensive comparisons, we implement two variants: (i) {\modelnameshort}-C-RN (base) and (ii) {\modelnameshort}-C-B32 (premium). 
The key distinction lies in the backbone of the CLIP image encoder.
Specifically, for {\modelnameshort}-C-RN, ResNet101 is employed on the VisDA dataset, while ResNet50 is used on the Office-Home and DomainNet-126 datasets. 
On the other hand, {\modelnameshort}-C-B32 adopts ViT-B/32~\cite{han2022survey} as the backbone across all datasets.
This extends to all other methods using CLIP. 
% Similarly, all methods using CLIP are marked with the same meaning for a convenience observation. 

\changedxx{
For an SFDA-focused comparison, we integrate the same CLIP model with existing SFDA methods, utilizing the CLIP's visual backbone (ViT B/32) as the feature extractor, followed by the original training algorithm. As such, all compared methods benefit from the CLIP pre-trained knowledge in a design-generic manner. 
Specifically, we select five updated SFDA methods as comparisons, including SHOT-B32, GKD-B32, NRC-B32, AaD-B32, and TPDS-B32. 
%Here, ``-32" means the adoption of ViT B/32 in CLIP, the same as {\modelnameshort}-C-B32.  
}

\textbf{Results.} 
% The quantitative comparisons are listed in Tab.~\ref{tab:oh-vc-sfda}$\sim$\ref{tab:126-sfda}.
As listed in Tab.~\ref{tab:oh-vc-sfda}$\sim$\ref{tab:126-sfda}, our {\modelnameshort}-C-RN and {\modelnameshort}-C-B32 perform best on all tasks across the three datasets except for two tasks. 
Compared with previous best SFDA methods TPDS~(on Office-Home), PLUE~(on VisDA) and GKD~(on DomainNet-126), {\modelnameshort}-C-RN improves by \textbf{4.1}\%, \textbf{1.0}\% and \textbf{9.3}\% in average accuracy, respectively. 
Compared with these UDA methods with both access to labelled source data and CLIP, {\modelnameshort}-C-RN still improves by \textbf{1.0}\%, \textbf{0.8}\%, and \textbf{2.8}\% on top of second-best methods PADCLIP-RN~(on Office-Home, VisDA) and ADCLIP-RN~(on DomainNet-126). 
The improvement expands further when we switch focus to the strong version {\modelnameshort}-C-B32. 
Such results are unsurprising since the well-pretrained CLIP is employed to identify the external causal elements.

On the other hand, {\modelnameshort}-C-RN defeats CLIP-RN on all tasks, achieving the improvement of \textbf{4.5}\%, \textbf{5.6}\%, and \textbf{5.3}\% on Office-Home, VisDA, and DomainNet-126 on average, respectively. 
Similarly, compared with CLIP-B32,  the corresponding improvement of {\modelnameshort}-C-B32 are changed to \textbf{5.1}\%, \textbf{5.1}\% and \textbf{3.0}\%.
This indicates the effect of the prompt learning to discover the external causal elements.

\changedxx{
Among the ``-B32" models, we have two observations. 
First, {\modelnameshort}-C-B32 outperforms all conventional methods across all tasks on the three datasets, confirming the effectiveness of our design.
Second, on the VisDA dataset, improvements with the B32 versions are limited, and they are even outperformed by the ResNet version, such as AaD and TPDS. 
This is because the rendered synthetic data (the source domain in VisDA) \changedrr{lacks} visual details, appearing primarily in shades of white and gray light and shadow.
When processed in ViT, this results in confused semantics.
As for AaD and TPDS, contrastive computation (AaD) and chain-like search (TPDS) amplify the noise in the confused semantics, leading to uncontrolled error propagation. 
}

\begin{table}[t]
    \centering
    \renewcommand\tabcolsep{3.2pt}
    \renewcommand\arraystretch{0.9}
    \scriptsize
    \caption{{Partial-set}, {Open-set} results (\%) on {\bf Office-Home}.}
        \begin{tabular}{llc|llc}
        \toprule
        \multicolumn{3}{c|}{Partial-set SFDA} &\multicolumn{3}{c}{Open-set SFDA}\\
        \midrule
        Method &Venue &{Avg.} &Method  &Venue &{Avg.} \\
        \midrule
        Source  &--  &{62.8} &Source 
        &--&{46.6} \\
        \midrule
        SHOT~\cite{liang2020we}        &ICML20     &79.3   &SHOT~\cite{liang2020we}        &ICML20 &{72.8} \\
        HCL~\cite{huang2021hcl}        &NeurIPS21  &79.6   &HCL~\cite{huang2021hcl}        &NeurIPS21 &{72.6} \\
        CoWA~\cite{lee2022confidence}  &ICML22     &83.2   &CoWA~\cite{lee2022confidence}  &ICML22 &{73.2} \\
        AaD~\cite{yang2022aad}         &NeurIPS22  &79.7   &AaD~\cite{yang2022aad}         &NeurIPS22 &{71.8} \\
        CRS~\cite{zhang2023crs}        &CVPR23     &80.6   &CRS~\cite{zhang2023crs}        &CVPR23 &{73.2} \\
        \midrule
        SHOT-B32~\cite{liang2020we}   &ICML20     &80.6   &SHOT-B32~\cite{liang2020we}  &ICML20     &73.3 \\
        AaD-B32~\cite{yang2022aad}    &NeurIPS22  &81.2   &AaD-B32~\cite{yang2022aad}   &NeurIPS22  &72.4 \\
        \rowcolor{gray! 40} \textbf{\modelnameshort}-C-RN  &--  &82.9 &\textbf{\modelnameshort}-C-RN &-- &79.6 \\
        \rowcolor{gray! 40} \textbf{\modelnameshort}-C-B32 &--  &\textbf{\color{cmblu}87.1} &\textbf{\modelnameshort}-C-B32 &-- &\textbf{\color{cmblu}84.0} \\
        % \rowcolor{gray! 40} \textbf{\modelnameshort}-C-B32 &--  &\textbf{\color{cmblu}85.8} &\textbf{\modelnameshort}-C-B32 &-- &\textbf{\color{cmblu}83.4} \\
        \bottomrule
    \end{tabular} 
    \label{tab:ps-os}
\end{table}

\subsection{Generalized SFDA} \label{sec:g-sfda}
\textbf{Dataset}. 
We evaluate {\modelnameshort} on Office-Home, following previous Generalized SFDA works~\cite{yang2021generalized,tang2023psat}.  

{\bf Evaluation protocol.}
Unlike \changedrr{Closed-set} SFDA, the Generalized SFDA problem highlights the anti-forgetting ability on the seen source domain. 
In terms of evaluation rule, the same as~\cite{yang2021generalized}, we adopt the harmonic mean accuracy that is computed by ${H}= ({2*A_s*A_t})/({A_s+A_t})$
where $A_s$ and $A_t$ are the accuracy on the source domain and the target domain, respectively.
Note that the $A_s$ is computed based on the source-testing set. 
The same \changedrr{as} \cite{yang2021generalized,tang2023psat}, on the source domain, the ratio of training and testing sets is 9:1.

{\bf Competitors.}
As listed in Tab.~\ref{tab:oh-st}, two Generalized SFDA comparisons, GDA~\cite{yang2021generalized} and PSAT~\cite{tang2023psat}, are additionally selected. 
GDA integrated domain attention with a local clustering-based self-supervised learning method. 
Essentially, knowledge of old tasks is preserved by constraining parameters near their original values.
Instead, PSAT enforces the adaptation process to remember the source domain by imposing source guidance, building a target domain-centric anti-forgetting mechanism. 
As {\modelnameshort}, we make two PSAT variants with ResNet50 (PSAT-RN) and ViT (PSAT-B32).

\textbf{One step adapting}.
Tab.~\ref{tab:oh-st} reports the results, including source and target domain accuracies and their average.
Among the ``-B32" methods, {\modelnameshort}-C-B32 (without any anti-forgetting design) defeats the best specially designed model PSAT-B32 (with an anti-forgetting mechanism) with a lead of {\bf 0.3}\% in harmonic mean accuracy.
{\modelnameshort}-C-RN outperforms other alternatives with the same backbone.
These results verify the competitiveness of our models in \changedrr{one-step adaptation}.

% An analysis on {continual adapting} is provided in \texttt{Appendix-\ref{app:continue-adap}}. 

% is only defeated by the best specially designed model PSAT-B32 (with anti-forgetting mechanism) with a gap of {\bf 0.4}\%.
% Tab.~\ref{tab:oh-st} shows that as one-step-adapting, {\modelnameshort}-C-B32 defeats the best specially designed model PSAT-B32 (with an anti-forgetting mechanism) with a lead of {\bf 0.3}\% in harmonic mean accuracy.  

\textbf{Continual adapting results.}
We conduct continual adaptation testing following~\cite{yang2021generalized}.
In the adaptation sequence Ar$\to$Cl$\to$Pr$\to$Rw, we first train an initial model on domain Ar, which then serves as the source for adaptation to domain Cl.
This process repeats iteratively until the final adaptation to domain Rw \changedrr{is achieved}, resulting in four distinct models.
Each model is evaluated across all domains, with {\bf 10}\% of the data in each domain randomly selected as the test set.
% The continual adaptation results are shown in the first sub-table of %Table~\ref{tab:cda-contin}, where the first column represents the adaptation sequence.

Tab.~\ref{tab:cda-contin} presents the results of {\modelnameshort}-C-RN and {\modelnameshort}-C-B32. 
We highlight two observations with {\modelnameshort}-C-B32:
{\bf First}, if a domain has been encountered during continual adaptation, the adapted model does not degrade significantly on that domain.
For example, in the adaptation sequence Cl$\to$Ar$\to$Pr$\to$Rw, when adapting from Cl to Ar, the accuracy on Ar reaches {\bf 83.4}\%. 
Subsequent adaptations to Pr and Rw lead to an average performance drop of {\bf 3.7}\% (see the second column in the second sub-table).
{\bf Second}, the adapted model performs well even on unseen domains.
For instance, along the adaptation path Ar$\to$Cl$\to$Pr$\to$Rw, the model’s accuracy on Rw gradually improves from {\bf 71.1}\% to {\bf 87.7}\% (see the fifth column in the first sub-table).
These results suggest that {\modelnameshort} effectively captures causal factors, making it robust to domain shifts.

\subsection{Open-set \& \changedrr{Partial-set} SFDA}
\textbf{Dataset}. We adopt Office-Home for evaluation~\cite{liang2020we,zhang2023crs}.

{\bf Competitors.}
Except Source, CLIP, SHOT, and CoWA as in the \changedrr{Closed-set} setting, we consider two more contrastive learning based SFDA methods, HCL~\cite{huang2021hcl} and CRS~\cite{zhang2023crs}.

% As presented in Tab.~\ref{tab:ps-os}, in terms of average accuracy, {\modelnameshort}-C-B32 surpasses not only all competitors with an improvement of {\bf 2.6}\% and {\bf 10.2}\% over CoWA (Partial-set) and CRS (Open-set). 

{\bf Comparison results.} 
As reported in Tab.~\ref{tab:ps-os}, {\modelnameshort}-C-B32 achieves the highest average accuracy, outperforming all competing methods by margins of {\bf 5.9}\% and {\bf 10.7}\% over AaD-B32 in the Partial-set setting and SHOT-B32 in the Open-set setting, respectively.
Similarly, it significantly outperforms SHOT-B32 and AaD-B32 under both Partial-set and Open-set settings, indicating the efficacy of the proposed design.
Only CoWA beats {\modelnameshort}-C-RN in the Partial-set setting with a tiny lead {\bf 0.3}\%.
The results indicate that {\modelnameshort} is competitive for the Partial-set and Open-set settings.

\subsection{Source-free out-of-distribution generalization} 
\textbf{Datasets}. 
We evaluate {\modelnameshort} in the SF-OODG setting on four challenging {\bf ImageNet Variants}.
Specifically, {\bf IN-V2} (i.e., ImageNet-V2)~\cite{imagenetV2} is an independent set consisting of natural images, collected from different sources, \changedrr{a total of} 10k images of 1000 ImageNet categories. 
{\bf IN-A} (i.e., ImageNet-A)~\cite{imagenetA} is a challenging set consisting of these "natural adversarial examples" (misclassified by a vanilla ResNet50~\cite{he2016deep}), containing 7.5k images of 200 ImageNet categories. 
{\bf IN-R}~(i.e., ImageNet-R)~\cite{imagenetR} is an ImageNet sub-set with artistic renditions. 
It contains 30k images covering 200 ImageNet categories. 
{\bf IN-K} (i.e., ImageNet-Sketch)~\cite{imagenetk} collects black and white sketches from the ImageNet validation set. This dataset contains 50k images from 1000 ImageNet categories.

{\bf Competitors.} 
In addition to the methods compared in the \changedrr{Closed-set} SFDA setting, we include four more: CoOP~\cite{zhou2022coop}, CoCoOp~\cite{zhou2022conditional}, TPT~\cite{shu2022test}, and ProGrad~\cite{zhu2023prompt}.
Among them, CoOP, CoCoOp, and ProGrad employ prompt learning with supervision from a few labeled samples, while TPT introduces self-supervised prompt tuning by minimizing multi-view entropy.

{\bf Comparison results}. 
Tab.~\ref{tab:ood-imagenet} reports the SF-OODG results.
In average accuracy, our methods, {\modelnameshort}-C-NR and {\modelnameshort}-C-B32, outperform previous SFDA methods by a significant margin, achieving at least a {\bf 10}\% improvement.
Even compared to CLIP and other CLIP-based methods, {\modelnameshort} maintains a clear advantage.
For example, {\modelnameshort}-C-NR improves upon CLIP-C-RN by {\bf 7.0}\% and surpasses the second-best CLIP-based method, TPT-RN, by {\bf 3.8}\%.
\changedxx{
Among the ``-B32" models, {\modelnameshort}-C-B32 excels with a margin of  at least {\bf 10.5}\% over previous methods.  
}

\subsection{Further analysis}

All experiments use the strong variant 
{\modelnameshort}-C-B32; the suffix “-C-B32” is omitted hereafter.

\subsubsection{Causal vs. statistical SFDA} \label{sec:cmp-difo-probe}

\changedrr{
In this strategic evaluation, we contrast \modelnameshort{} with two state-of-the-art statistical SFDA methods, DIFO \cite{tang2024source} and ProDe \cite{tang2025proxy}, both of which are equipped with the same ViL model. Our analysis focuses on two critical modeling aspects, examined across various SFDA settings:
{\it Domain fitness}: Evaluated under Closed-set conditions; 
{\it Domain generalization}: Assessed across Generalized, Partial-set, Open-set, and SF-OODG settings.
% , and through noise resilience.
}

Based on this evaluation, we draw the following key observations:
(1) Domain fitness (Closed-set SFDA): As presented in Tab.~\ref{tab:comp-difo-prode}, \modelnameshort{} is indeed outperformed by both DIFO and ProDe in the Closed-set setting. This is attributed to the inherent statistical learning nature of these alternatives, which tends to overfit their models towards the specific target domain.
(2) Under the Generalized SFDA setting, {\modelnameshort} demonstrates superior performance in H-accuracy, also exhibiting the least drop in source domain accuracy (see S-accuracy). 
This is because the captured causality mitigates the risk of forgetting the source domain.
%
% (3) Broad domain generalization across the Open-set, Partial-set, and SF-OODG settings, 
% \modelnameshort{} consistently stands out with strong performance. 
(3) Except for the Closed-set scenario, \modelnameshort{} consistently stands out with strong performance on the other four settings. 
This highlights the superior domain generalization capabilities achieved through our method's causal knowledge discovery. Importantly, our approach provides the best overall solution across all these diverse settings.
%
% (4) Anti-noise capability: As illustrated in Fig.~\ref{fig:noise-difo-prode}, \modelnameshort{} consistently proves to be the most robust and sustainable method across a wide range of noise degrees. This underscores its excellent anti-noise capability.
% %
Collectively, these results validate that our approach strikes the optimal balance between domain fitness and generalization, thereby laying a solid foundation to serve as a unified SFDA solution.

\begin{table}[t]
    \caption{SF-OODG results (\%) on {\bf ImageNet variants}. 
    \changed{{L}, {ViL} means label-required, ViL model-required, respectively.}
    }
    \label{tab:ood-imagenet}
    \scriptsize
    \centering
    % \footnotesize
    \renewcommand\tabcolsep{3pt}
    \renewcommand\arraystretch{0.9}
    % \begin{tabular}{*{9}{c}}
    \begin{tabular}{l l c c  c c c c c } 
    \toprule
    \multirow{3}{*}{Method} &\multirow{3}{*}{Venue} &\multirow{3}{*}{L} &\multirow{3}{*}{ViL} &\multicolumn{5}{c}{{\bf ImageNet}$\to$X} \\
    \cmidrule(lr){5-9}
    &  &  &&{\bf IN-V2} &{\bf IN-K} &{\bf IN-A} &{\bf IN-R} & {Avg.}\\
    \midrule
    Source     &--     &-- &--        &62.7  &22.2  &0.7   &35.1  &30.2 \\
    CLIP-RN~\cite{radford2021learning}    &ICML21 &-- &\cmark    &51.5  &33.3 	&21.8  &56.1  &40.7 \\
    CLIP-B32~\cite{radford2021learning}   &ICML21 &-- &\cmark    &54.8  &40.8  &29.5  &66.2  &47.8 \\
    \midrule
    CoOP-RN~\cite{zhou2022coop}      &IJCV21 &\cmark &\cmark &55.4 &34.7 &23.1 &56.6 &42.4 \\
    CoOP-B32~\cite{zhou2022coop}     &IJCV21 &\cmark &\cmark &58.2 &41.5 &31.3 &65.8 &49.2 \\
    CoCoOp-RN~\cite{zhou2022conditional}    &CVPR22 &\cmark &\cmark &55.7 &34.5 &23.3 &57.7 &42.8 \\
    CoCoOp-B32~\cite{zhou2022conditional}   &CVPR22 &\cmark &\cmark &56.6 &40.7 &30.3 &64.1 &47.9 \\
    TPT-RN~\cite{shu2022test}               &NeurIPS22 &\xmark &\cmark &54.7 &35.1 &26.7 &59.1 &43.9 \\
    ProGrad-RN~\cite{zhu2023prompt}         &ICCV24 &\cmark &\cmark &54.7 &34.4 &23.1 &56.8 &42.2 \\
    ProGrad-B32~\cite{zhu2023prompt}        &ICCV24 &\cmark &\cmark &57.4 &41.7 &\textbf{\color{cmblu}31.9} &66.5 &49.4 \\
    % CoCoOP &CVPR22 &\cmark &-- &0.0 &0.0 &0.0 &0.0 \\	
    \midrule
    SHOT~\cite{liang2020we}            &ICML20 &\xmark &\xmark    &62.5 &38.4 &2.1 &42.7 &36.4 \\
    GKD~\cite{tang2021model}           &IROS21 &\xmark &\xmark    &62.3 &38.2 &2.1 &42.2 &36.2 \\
    NRC~\cite{yang2021nrc}             &NeurIPS21 &\xmark &\xmark    &61.5 &0.9  &1.3 &28.9 &23.1 \\
    AdaCon~\cite{chen2022contrastive}  &CVPR22 &\xmark &\xmark    &50.9 &18.5 &2.0 &38.6 &27.5 \\
    CoWA~\cite{lee2022confidence}      &ICML22 &\xmark &\xmark    &62.7 &37.8 &1.7 &46.3 &37.1 \\
    PLUE~\cite{litrico2023guiding}     &CVPR23 &\xmark &\xmark    &53.0 &18.5 &2.0 &38.6 &28.0 \\
    TPDS~\cite{tang2023tpds}           &IJCV24 &\xmark &\xmark    &62.9 &35.0 &2.9 &48.4 &37.3 \\
    % TPDS       &IJCV23 &\xmark    &0.0 &0.0 &0.0 &0.0 &0.0  &0.0 \\ 
    \midrule
    SHOT-B32~\cite{liang2020we}     &ICML20      &\xmark &\cmark    &\textbf{\color{cmblu}69.4} & 43.6 & 1.6 & 58.2 & 43.2 \\
    GKD-B32~\cite{tang2021model}    &IROS21      &\xmark &\cmark    &53.5 & 45.2 & 1.0 & 58.3 & 39.5 \\
    NRC-B32~\cite{yang2021nrc}      &NeurIPS21   &\xmark &\cmark    &66.2 & 18.6 & 15.4 & 38.0 & 34.6 \\
    TPDS-B32~\cite{tang2023tpds}    &IJCV24      &\xmark &\cmark    &69.0 & 38.9 & 0.6 & 58.7 & 41.8 \\
    % DIFO-B32~\cite{tang2024source}   &CVPR2024      &\xmark &\cmark &59.6 &37.7 &25.2 &66.4 &47.2\\
    % ProDe-B32~\cite{tang2024proxy}   &ICLR2025      &\xmark &\cmark &62.1 &45.8 &24.7 &72.5 &51.3\\   

    \rowcolor{gray! 40}  {\bf {\modelnameshort}}-C-RN &-- &\xmark &\cmark    &64.4 &42.6 &22.3 &61.5 &47.7 \\
    \rowcolor{gray! 40}  {\bf {\modelnameshort}}-C-B32 &-- &\xmark &\cmark   &64.7 &\textbf{\color{cmblu}48.4} &30.6 &\textbf{\color{cmblu}71.0} &\textbf{\color{cmblu}53.7} \\
    \bottomrule
    \end{tabular}
\end{table}

\begin{table*}[t]
    \caption{
    \changedrr{Causal vs. statistical SFDA. 
    %, DIFO and ProDe on all five settings grouped by Closed-set setting and Extended setting. 
    OH/VDA/DN126: {\bf Office-Home}/{\bf VisDA}/{\bf DomainNet-126}. 
    {S}/{T}: Source/Target domain;
    $H$: Harmonic mean accuracy.
    %and {H} are the results on the source and target domains, and harmonic mean accuracy, respectively.
    % \textbf{\color{Plum}98.1} is the average accuracy of the source model on Office-Home. 
    % ${{H}_{all}}$: Average over 
    % all settings ($H$ metric used for Generalized SFDA).
    %by averaging {\it Avg.} of closed-set, H-accuracy in Generalized SFDA, {\it Avg.} in SF-OODG, and accuracy of Open-set and Partial-set.
    ${H}_{g}$: Average over 
    all settings except closed-set.
    %evaluates the domain generalization ability, excluding Closed-set results.
    }
    }
	\label{tab:comp-difo-prode}	
	\renewcommand\tabcolsep{5.0pt}
	\renewcommand\arraystretch{0.9}
	\scriptsize
	\centering
 	\begin{tabular}{l l c c c c | c c c  c c  c c c c c c}
        \toprule
        \multirow{3}{*}{Method} &\multirow{3}{*}{Venue} &\multicolumn{4}{c}{\bf Closed-set} \vline &\multicolumn{3}{c}{{\bf Generalized}}  &\multirow{1}{*}{\bf Open-set} &\multirow{1}{*}{\bf Partial-set} &\multicolumn{5}{c}{\bf SF-OODG} & \\
        \cmidrule(r){3-6} \cmidrule(r){7-9} \cmidrule(r){10-10} \cmidrule(r){11-11} \cmidrule(r){12-16}
        & &OH &VDA &DN126 &{\it Avg}. &S &T & $H$ &T &T &{IN-V2} &{IN-K} &{IN-A} &{IN-R} &{\it Avg.} &${{H}_{g}}$ \\
        \midrule
        Source &-- &59.2 &49.2 &54.7 &54.4 &98.1 &59.2 &73.8 &46.6 &62.8 &62.7 &22.2 &0.7 &35.1 &30.2 &53.4 \\
        \midrule
        DIFO~\cite{tang2024source}  &CVPR24    &83.1 &90.3 &80.0  &84.5 &78.0  &83.1  &80.5 &75.9 &85.6 &59.6 &37.7 &25.2 &66.4 &47.2 &72.3 \\
        ProDe~\cite{tang2024proxy}  &ICLR25    &\textbf{\color{cmblu}84.5} &\textbf{\color{cmblu}91.0} &\textbf{\color{cmblu}85.0}  &\textbf{\color{cmblu}86.8} &85.1 &84.5  &84.8  &82.6 &84.2 &62.1 &45.8 &24.7 &69.7 &50.6 &75.5 \\         
        \rowcolor{gray! 30} {\bf {\modelnameshort}}  &--  &84.3 &90.3 &82.2  &85.6 &86.1 &84.3 &\textbf{\color{cmblu}85.2} &\textbf{\color{cmblu}84.0} &\textbf{\color{cmblu}87.1}  &\textbf{\color{cmblu}64.7} &\textbf{\color{cmblu}48.4} &\textbf{\color{cmblu}30.6} &\textbf{\color{cmblu}71.0} 
        &\textbf{\color{cmblu}53.7} &\textbf{\color{cmblu}77.5}\\
        \bottomrule
  	\end{tabular}
\end{table*}
\begin{table*}[t]
	\caption{
        \changedrrr{Unification comparison results (\%). 
        $H_{all}$: Average over all settings; $H_{wrg}$: The largest percentage shortfall of a method from the per-setting best across all settings; $H_{loso}$: Omit one setting at a time and average the scores over the rest.}
        }
	\label{tab:unification-analysis}
	\renewcommand\tabcolsep{8.0pt}
	\renewcommand\arraystretch{0.95}
	\scriptsize
	\centering
 	\begin{tabular}{ l l c c c c c c}
        \toprule
        \multirow{2}{*}{Method}   &\multirow{2}{*}{$H_{wrg}$($\downarrow$)} &\multicolumn{5}{c}{$H_{loso}$($\uparrow$)} &\multirow{2}{*}{$H_{all}$($\uparrow$)}\\
        \cmidrule(r){3-7}
        &  &w/o Closed-set &w/o Generalized setting &w/o Open-set &w/o Partial-set &w/o SF-OODG &\\
        \midrule
        % DIFO~\cite{tang2024source}       &12.1  &72.3  &73.3  &74.4 &72.0  &81.6 & 74.7\\
        % ProDe~\cite{tang2024proxy}       &5.77  &75.5  &76.0  &76.6 &76.2  &\textbf{\color{cmblu}84.6} & 77.7\\     
        % \rowcolor{gray! 40} {\bf {\modelnameshort}}     &\textbf{\color{cmblu}2.99}  &\textbf{\color{cmblu}76.9}  &\textbf{\color{cmblu}76.8}  &\textbf{\color{cmblu}77.1} &\textbf{\color{cmblu}76.5}     &84.5 &\textbf{\color{cmblu}78.4}\\

        DIFO~\cite{tang2024source}       &12.1  &72.3  &73.3  &74.4 &72.0  &81.6 & 74.7\\
        ProDe~\cite{tang2024proxy}       &5.77  &75.5  &76.0  &76.6 &76.2  &84.6 & 77.7\\     
        \rowcolor{gray! 30} {\bf {\modelnameshort}}      &\textbf{\color{cmblu}1.38}  &\textbf{\color{cmblu}77.5}  &\textbf{\color{cmblu}77.6}  &\textbf{\color{cmblu}77.9} &\textbf{\color{cmblu}77.1}  &\textbf{\color{cmblu}85.5} &\textbf{\color{cmblu}79.1}\\
        
        \bottomrule
  	\end{tabular}
\end{table*}

{\bf Unification comparison.}
To underscore \modelnameshort's advantage over DIFO and ProDe, we perform a unification evaluation using the metrics in Eq.~(\ref{eqn:unification-metrics}). 
The comparison in Tab.~\ref{tab:unification-analysis} shows that {\modelnameshort} attains the highest score in all scenarios.
% , leading to at least \textbf{xx\%} improvement in overall average $H_{\text{all}}$.
% 
% As shown in Tab.~\ref{tab:unification-analysis}, {\modelnameshort} attains the highest score in every scenario except one: When SF-OODG is omitted in $H_{\mathrm{loso}}$, where the gap is only \textbf{0.1\%}. 
% 
This indicates that {\modelnameshort} delivers robust, stable performance across diverse settings. Moreover, it exhibits no apparent weakness in any single scenario (cf.\ $H_{\mathrm{wrg}}$), and its all-round superiority is not driven by any individual setting (cf.\ $H_{\mathrm{loso}}$). Together, this evidence supports our claim of being towards a universally adaptable framework. 

% \changedrrr{
% {\bf Unification comparison.}
% To underscore \modelnameshort's advantage over DIFO and ProDe, we perform a unification evaluation using the metrics in Eq.~(\ref{eqn:unification-metrics}). As shown in Tab.~\ref{tab:unification-analysis}, {\modelnameshort} attains the highest score in every scenario except one: When SF-OODG is omitted in $H_{\mathrm{loso}}$, where the gap is only \textbf{0.1\%}. These results indicate that {\modelnameshort} delivers robust, stable performance across diverse settings. Moreover, it exhibits no apparent weakness in any single scenario (cf.\ $H_{\mathrm{wrg}}$), and its all-round superiority is not driven by any individual setting (cf.\ $H_{\mathrm{loso}}$). Together, this evidence supports our claim of being a universally adaptable framework. 
% }

\subsubsection{Analysis for causal factors discovery} \label{sec:cau-ana}
We conduct causality analysis in two aspects: (1) complementarity of internal and external causality, and (2) analysis for the discovery process based on pseudo-label.
% % (3) Robustness to noise.

\begin{figure}[t]
    \begin{center}
        \subfigure{
            \includegraphics[width=1.0\linewidth]{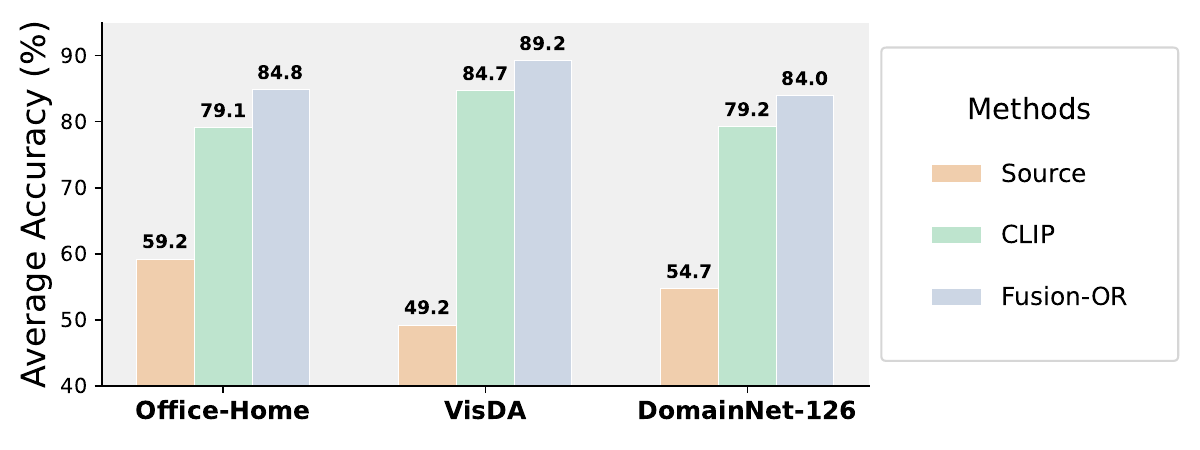}}
    \end{center}
    \setlength{\abovecaptionskip}{0cm}
    \caption{
    \changedxx{
    Zero-shot fusion results (Fusion-OR) of CLIP and the source model (Source) under the \changedrr{Closed-set} SFDA setting on the {\bf Office-Home}, {\bf VisDA}, and {\bf  DomainNet-126} datasets. This fusion method follows the logical OR principle: An input instance is considered correctly classified if either CLIP or the source model predicts its ground truth category.
    }    
    }
    \label{fig:mix-or}
\end{figure}

\changed{
{\bf Complementarity of internal and external causality.} 
As aforementioned, causal factors originate from both the domain data itself (encoded internal causality) and external knowledge sources (encoded external causality). 
To illustrate this complementarity, we conduct an intuitive empirical experiment on the target domain. Specifically, we fuse the zero-shot predictions of the source model (Source) and a ViL model like CLIP using a logical OR operation. That is, if either CLIP or Source correctly predicts the true category of an input instance $\boldsymbol{x}$, we consider $\boldsymbol{x}$ correctly classified. As shown in Fig.~\ref{fig:mix-or}, Fusion-OR significantly outperforms both Source and CLIP individually, reinforcing the plausibility of our internal-external causality partition.
}

\begin{figure}[t]
   \setlength{\abovecaptionskip}{0cm}
   \centering  
   \subfigure{\includegraphics[width=0.77\linewidth]{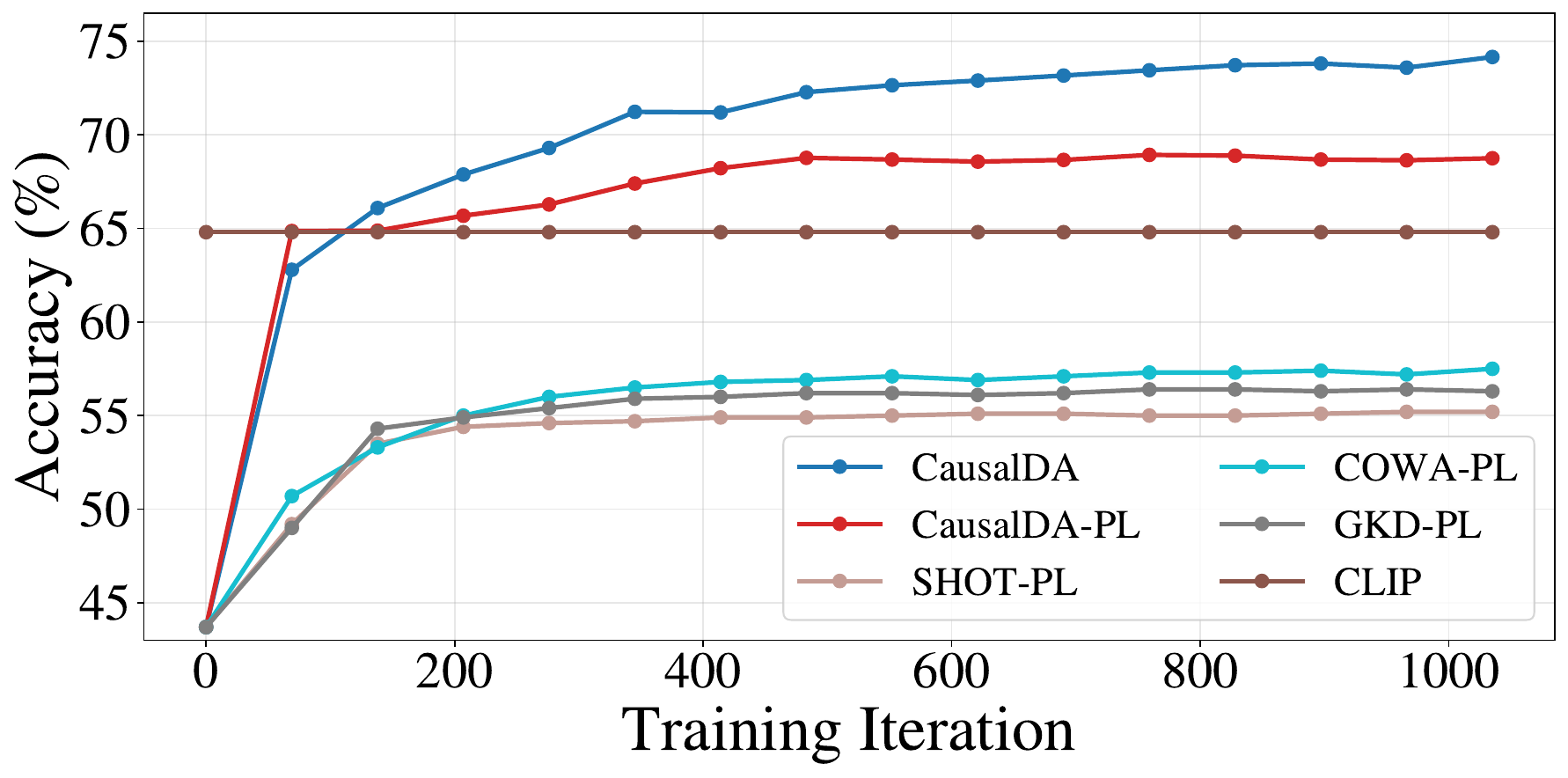}} 
   \caption{Dynamics of pseudo-label accuracy on task Ar$\rightarrow$Cl in the {\bf Office-home} dataset. 
    The method "X-PL" means the corresponding accuracy of pseudo-label generated in X.}
   \label{fig:pl-comp}  
\end{figure}

\textbf{Pseudo-label-based analysis for discovery process}.
In {\modelnameshort}, the external causal elements $S_e$ are identified and converted into pseudo-labels.
Thus, the quality of these pseudo-labels serves as an indicator of the discovery process.
To evaluate this, we compare pseudo-label-based methods SHOT, COWA, and GKD, along with CLIP’s zero-shot predictions (denoted as CLIP).
The results, shown in Fig.~\ref{fig:pl-comp}, demonstrate that {\modelnameshort}-PL achieves significantly higher pseudo-label accuracy than prior methods, benefiting from CLIP-assisted $S_e$ discovery.
Furthermore, the performance curve of {\modelnameshort} remains consistently above that of {\modelnameshort}-PL, with a significant gap throughout training.
This highlights the effectiveness of capturing internal causal elements $S_i$, further enhancing model performance.

\vspace{5pt}
\subsubsection{Model analysis} \label{sec:mod-ana}
\begin{figure*}[t]
	\setlength{\abovecaptionskip}{-0.1cm}
	\begin{center}
		\subfigure{
			\includegraphics[width=0.14\linewidth,height=0.13\linewidth]{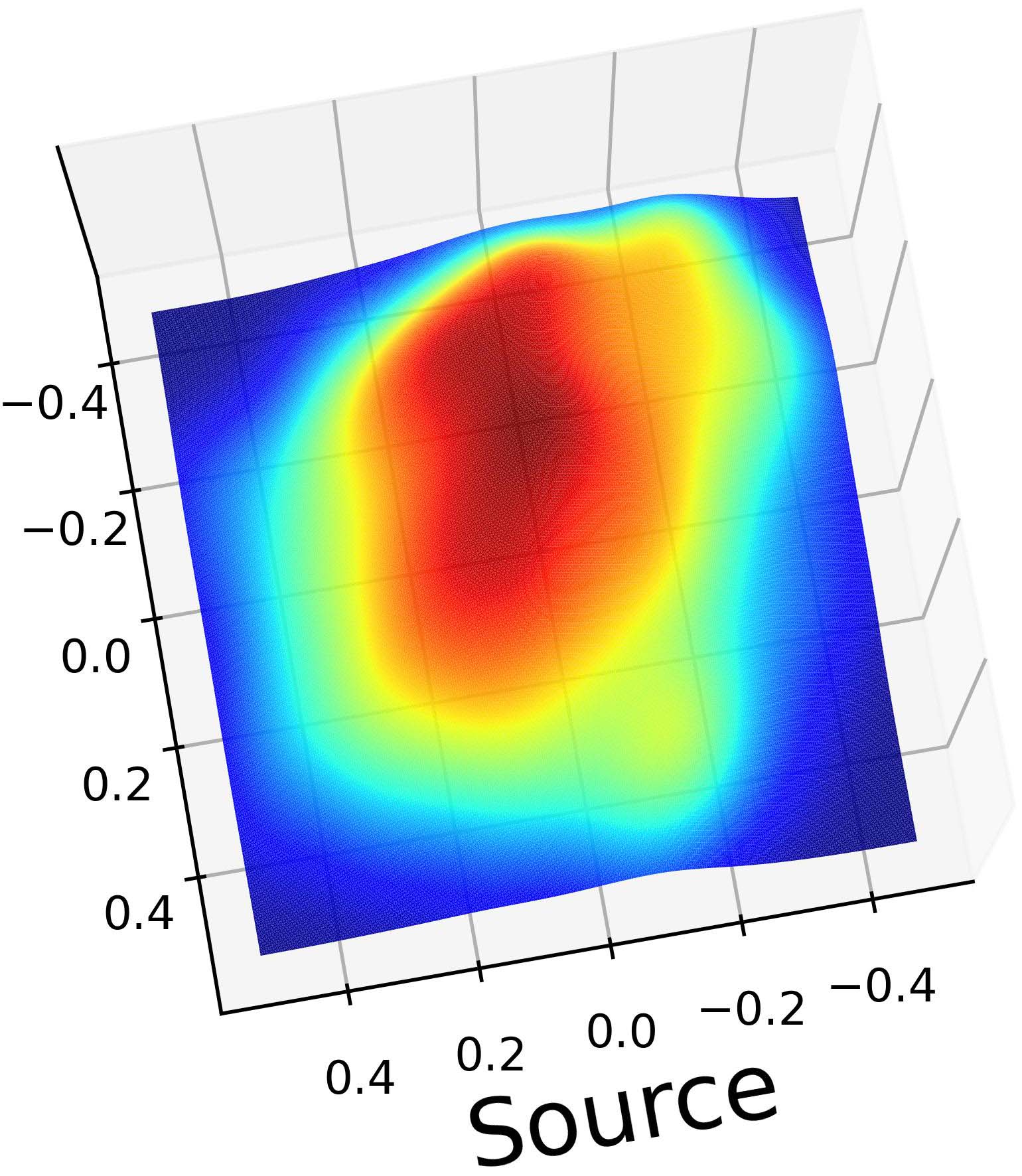}}
		\subfigure{
			\includegraphics[width=0.14\linewidth,height=0.13\linewidth]{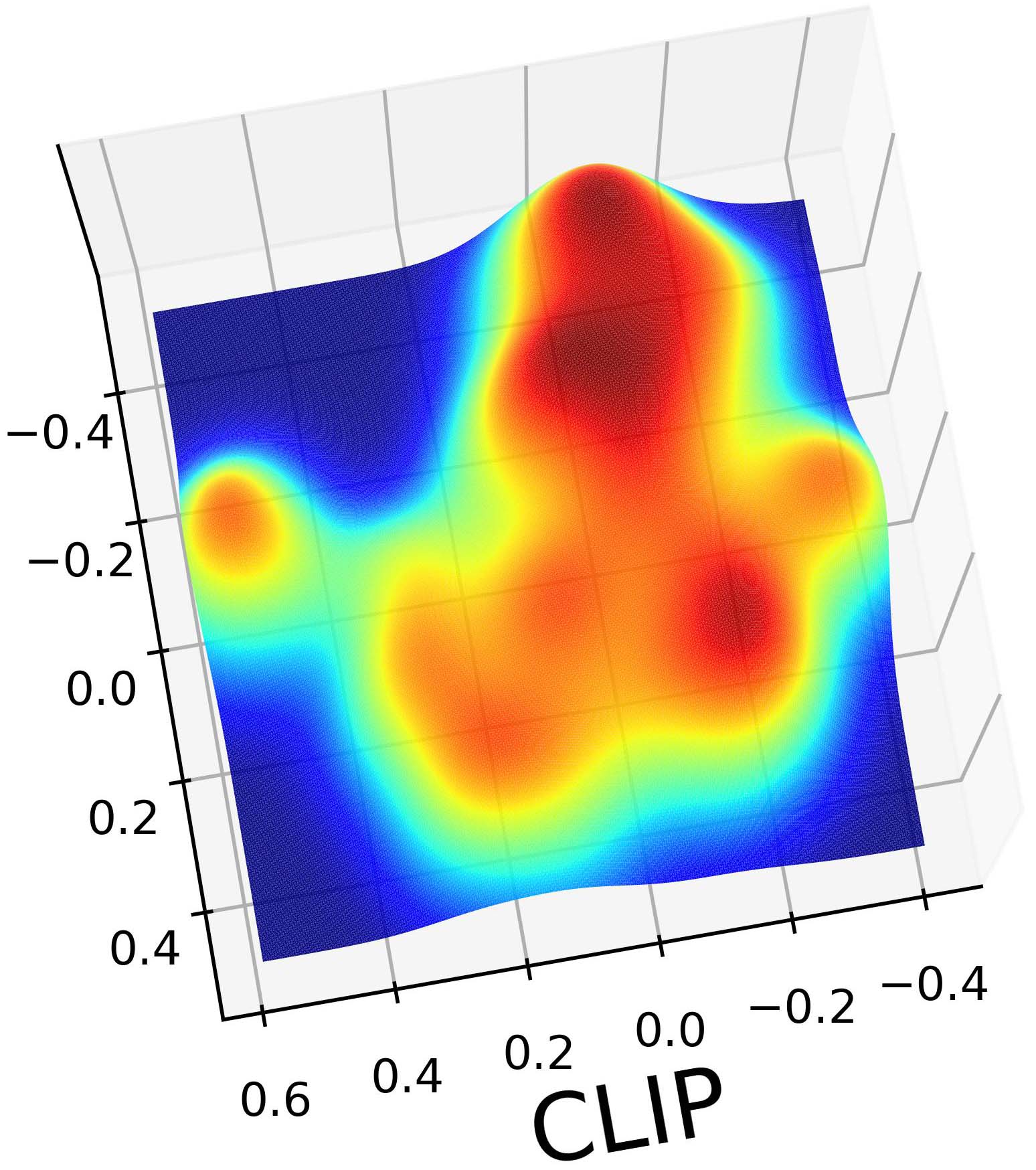}}
            \subfigure{
			\includegraphics[width=0.14\linewidth,height=0.13\linewidth]{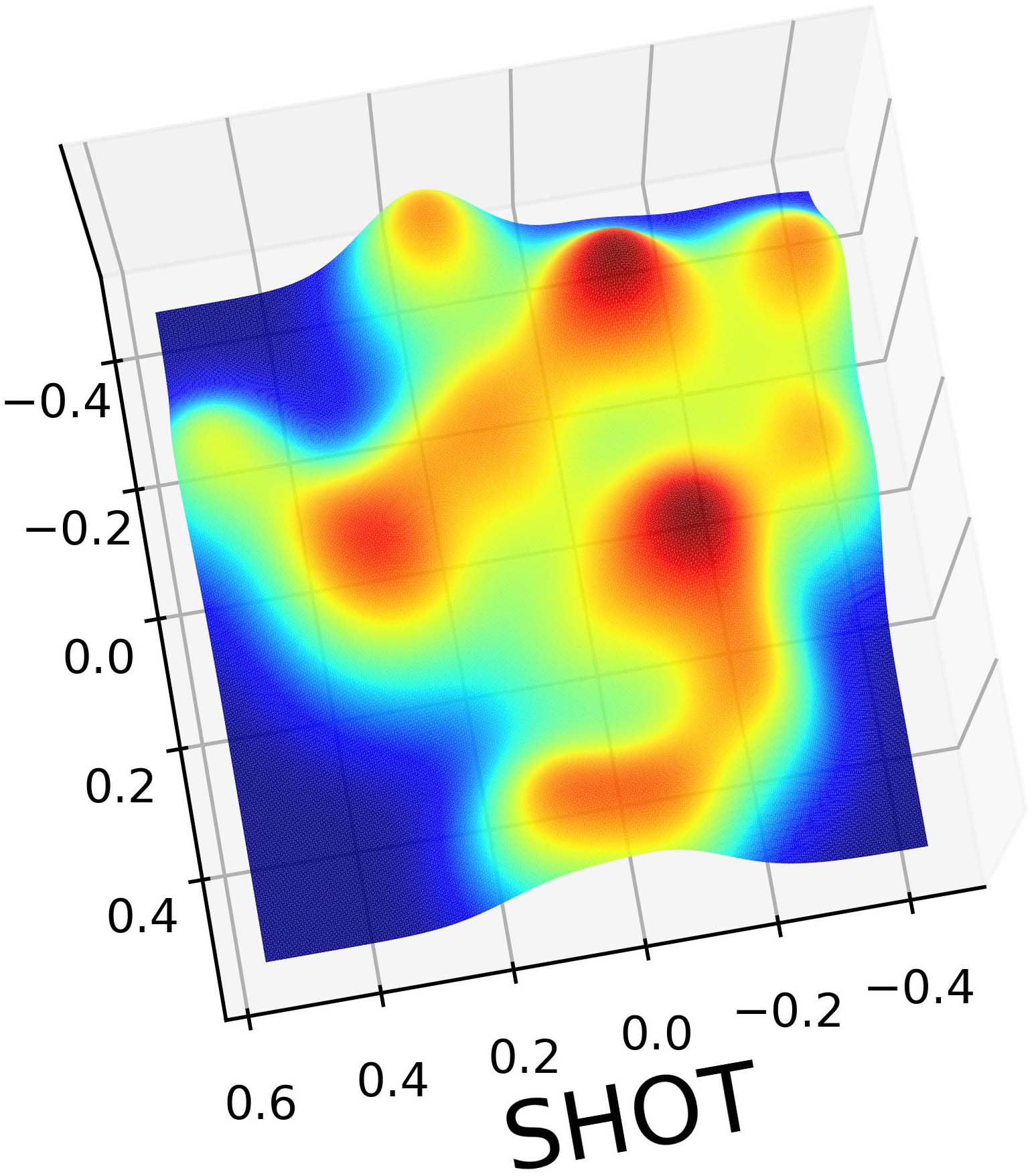}}
            \subfigure{
    		\includegraphics[width=0.14\linewidth,height=0.13\linewidth]{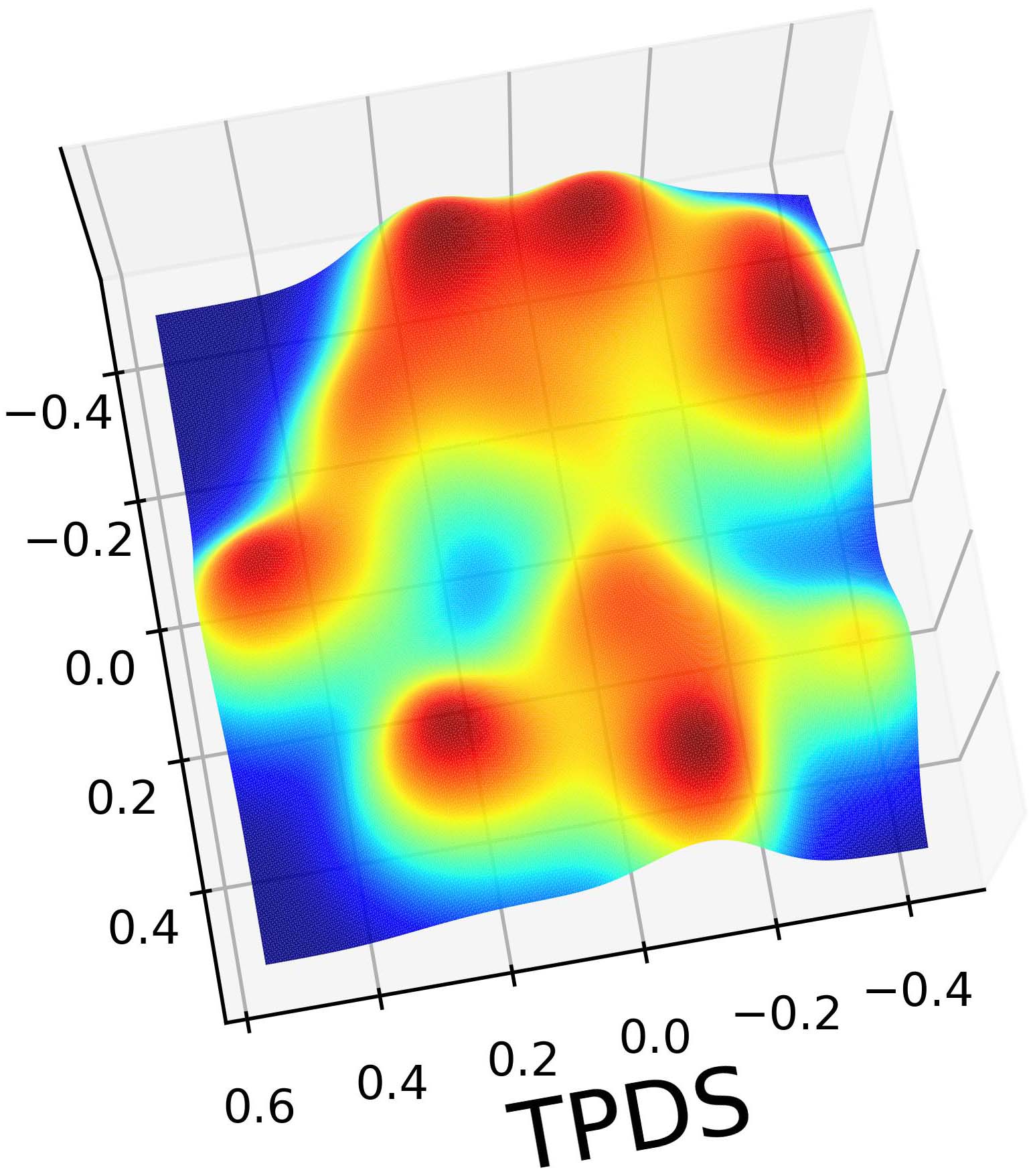}}
            \subfigure{
    		\includegraphics[width=0.14\linewidth,height=0.13\linewidth]{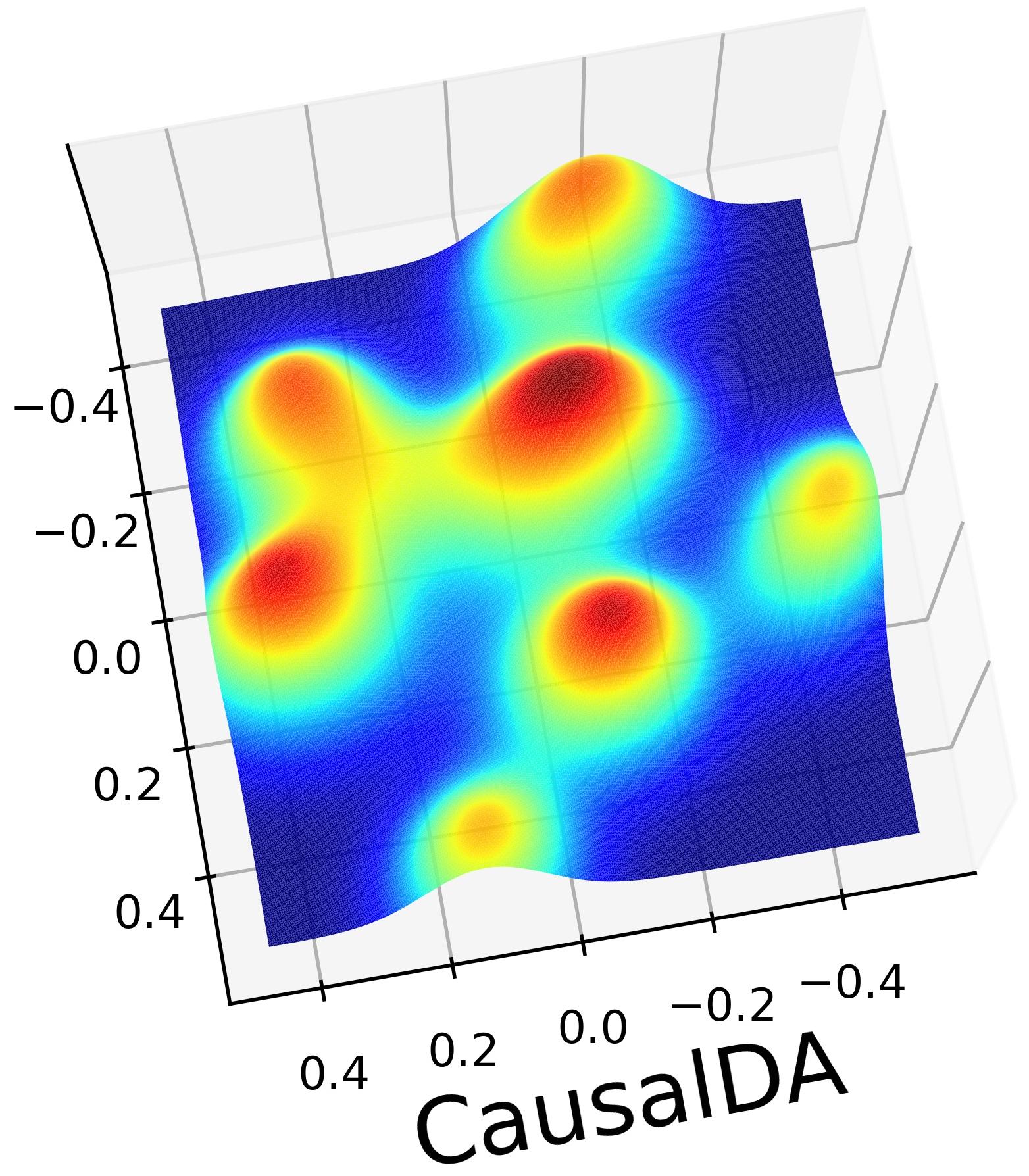}}
            \subfigure{
    		\includegraphics[width=0.14\linewidth,height=0.13\linewidth]{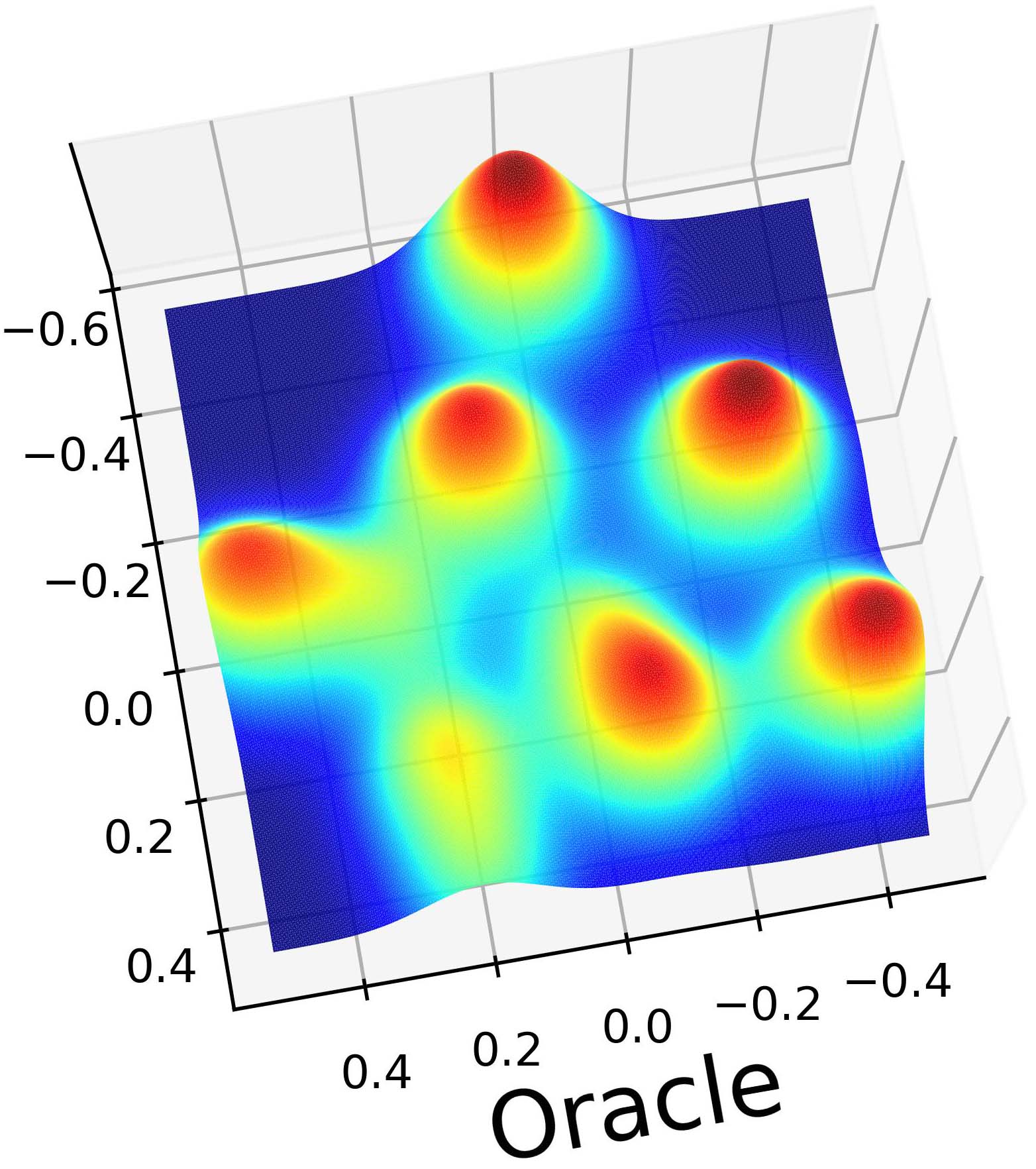}}
	\end{center}
	\caption{Logit distribution visualization on transfer task Ar$\to$Cl in {\bf Office-Home}. Oracle: Model trained on the real labels of domain Cl. For a clear view, the first ten categories are displayed. 
    } 
    \label{fig:tsne}
\end{figure*}

\textbf{Feature distribution visualization}.
Feature distribution provides an intuitive way to evaluate classification performance.
We compare with Source (source model), CLIP, SHOT, TPDS, and Oracle (trained on domain Cl with real labels).
Fig.~\ref{fig:tsne} presents the logit visualization results using a 3D density chart.
As we move from Source to {\modelnameshort}, the clustering effect progressively improves, with {\modelnameshort} exhibiting the most structured distribution, closely resembling Oracle.

\textbf{Ablation study}.
In {\modelnameshort}, we separately discover the causal factors $S\!=\!\{S_e,S_i\}$. To isolate the effect of this strategy, we evaluate \changed{four} variations:
(i) {{\modelnameshort}-w-$S_i$}:
We use only $\mathcal{L_{\rm{UN}}}$ from Eq.~\eqref{eqn:loss-ssl-ins} as the objective for model training, allowing us to assess the impact of capturing $S_i$.
(ii) {{\modelnameshort}-w-$S_e$}:
We apply pseudo-label supervised learning, regulated by $\mathcal{L}_{\rm{SCE}}$, where the pseudo-labels are generated from $S_e$ discovery, regulated by 
$L_{\rm{EC}}$. This isolates the effect of capturing $S_e$.
(iii) {{\modelnameshort}-w-CLIP}:
We directly use CLIP’s zero-shot results as pseudo-labels to regulate the discovery of 
$S_i$.
\changed{(iv) {{\modelnameshort}-w-P1}:
We retain only the phase-1 objective, removing the loss $\mathcal{L_{\rm{IC}}}$ from Eq.~\eqref{eqn:loss-ssl-ins}.}

Tab.~\ref{tab:abl-exp} presents the results of these methods across four SFDA settings and SF-OODG.
Specifically, both {{\modelnameshort}-w-$S_i$} and {{\modelnameshort}-w-$S_e$} outperform the Source model, but fall significantly behind the full {\modelnameshort} version.
This suggests that capturing either $S_e$ or 
$S_i$ independently is effective, but both are necessary for optimal performance.
The comparison between {{\modelnameshort}-w-CLIP} and {\modelnameshort} highlights the importance of $L_{\rm{EC}}$-based prompt learning for $S_e$ discovery.
\changed{Finally, {{\modelnameshort}-w-P1} performs considerably worse than {\modelnameshort}, emphasizing the value of our two-stage design in the causal factor capture pipeline.}

% width=0.29\linewidth,height=0.30\linewidth

\textbf{Scalability study of IC instantiation}.
The objective \( \mathcal{L}_{\rm{IC}} \) in Eq.~\eqref{eqn:loss-ssl-ins}, which regulates \( S_i \) discovery, is a component with multiple implementation options.  
To evaluate its scalability, we present three alternative instantiations for the two regulators in \( \mathcal{L}_{\rm{IC}} \).
For \( \mathcal{L}_{\rm{UN}} \), we consider the following alternatives:  
(i) {{\modelnameshort}+ENT}: Entropy regularization without category balance constraints,  
(ii) {{\modelnameshort}+NRC}: Data local structure-based regularization,  
(iii) {{\modelnameshort}+TPDS}: Domain shift control-based regularization.
For \( \mathcal{L}_{\rm{SCE}} \), we test three popular methods: 1-Norm (\( ||\cdot||_1 \)), 2-Norm (\( ||\cdot||_2 \)), and KL Divergence.

As shown in Tab.~\ref{tab:extend-exp}, in the Closed-set setting, {{\modelnameshort}+TPDS} achieves the best results, with other methods lagging by up to {\bf 1.9}\% (on VisDA). 
This indicates that these typical regulators do not significantly affect model performance in this setting. 
In the SF-OODG setting, all methods (except {{\modelnameshort}+KL}) show a slight accuracy drop compared to {\modelnameshort}.  
For {{\modelnameshort}+ENT}, removing the category balance regularization leads to a performance decrease.  
The performance loss for {{\modelnameshort}+NRC} and {{\modelnameshort}+TPDS} is attributed to the fact that these methods do not specifically address the SF-OODG setting.

For \( \mathcal{L}_{\rm{SCE}} \), the results show that probability distribution approximation (KL Divergence) outperforms feature-based methods (\( ||\cdot||_1 \) and \( ||\cdot||_2 \)) in preserving model performance in the presence of external causal elements.

\begin{table}[t]
\caption{Ablation study results (\%). 
CS, OS, PS, and GS mean the Closed-set, Open-set, Partial-set, and \changedrr{Generalized} SFDA settings, respectively.}
    \label{tab:abl-exp}
    \renewcommand\tabcolsep{6.3pt}
    \renewcommand\arraystretch{0.95} 
    % \footnotesize
    \scriptsize
    \centering
    \begin{tabular}{ l c c c c c c }
        \toprule
        \multirow{3}{*}{Method}  &\multicolumn{5}{c}{\bf Office-Home}  &{\bf IN-K} \\
        \cmidrule(lr){2-6} \cmidrule(lr){7-7}
        &{CS}  &{OS} &{PS} &{GS} &Avg. &{SF-OODG} \\
        \midrule
        % Source                            &59.2  &46.6  &62.8  &73.1  &60.4  &30.2 \\
        % {\bf {\modelnameshort}}-w-$S_i$   &70.3  &71.5  &70.1  &75.9  &72.6  &36.9 \\
        % {\bf {\modelnameshort}}-w-$S_e$   &81.6  &80.7  &84.1  &84.3  &82.2  &44.2 \\
        % {\bf {\modelnameshort}}-w-CLIP    &80.9  &78.3  &83.9  &82.5  &80.6  &44.2 \\
        % {\bf {\modelnameshort}}-w-P1      &79.3  &37.3  &79.3  &77.2  &68.3  &40.8 \\
        % \rowcolor{gray! 40} {\bf {\modelnameshort}}  &\textbf{\color{cmblu}83.3}  &\textbf{\color{cmblu}83.4}     &\textbf{\color{cmblu}85.8}&\textbf{\color{cmblu}84.5} &\textbf{\color{cmblu}83.7} &\textbf{\color{cmblu}48.4} \\ 
        Source                            &59.2  &46.6  &62.8  &73.1  &60.4  &30.2 \\
        {\bf {\modelnameshort}}-w-$S_i$   &71.3  &72.1  &71.6  &76.6  &72.9  &36.9 \\
        {\bf {\modelnameshort}}-w-$S_e$   &82.5  &81.4  &85.5  &84.9  &83.6  &44.2 \\
        {\bf {\modelnameshort}}-w-CLIP    &82.0  &78.8  &85.1  &83.3  &82.3  &44.2 \\
        {\bf {\modelnameshort}}-w-P1      &80.3  &38.1  &80.7  &78.1  &69.3  &40.8 \\
        \rowcolor{gray! 40} {\bf {\modelnameshort}}  &\textbf{\color{cmblu}84.3}  &\textbf{\color{cmblu}84.0}     &\textbf{\color{cmblu}87.1}&\textbf{\color{cmblu}85.2} &\textbf{\color{cmblu}85.1} &\textbf{\color{cmblu}48.4} \\
        \bottomrule
    \end{tabular}
\end{table}

\begin{table}[t]
    \caption{Scalability study~(\%) of IC instantiation.} 
    \label{tab:scalability}
    \renewcommand\tabcolsep{7.0pt}
    \renewcommand\arraystretch{0.95}
    % \footnotesize
    \scriptsize
    \centering
    \begin{tabular}{ c l c c c}
        \toprule
        % \multirow{2}{*}{Method} &\multicolumn{13}{c}{\textbf{Office-Home}} \vline &{\textbf{VisDA}} \\
        \multirow{3}{*}{Loss} &\multirow{3}{*}{Method}  &\multicolumn{2}{c}{Closed-set}  &{SF-OODG} \\
        \cmidrule(lr){3-4} \cmidrule(lr){5-5}
        & &{\bf Office-Home} &{\bf VisDA} &{\bf IN-K} \\
        \midrule
        % \multirow{3}{*}{$\cal{L}_{\rm{UN}}$} 
        % &{\bf {\modelnameshort}}+ENT   &82.5  &89.4   &46.8 \\
        % &{\bf {\modelnameshort}}+NRC   &83.0  &87.9   &47.5 \\
        % &{\bf {\modelnameshort}}+TPDS  &\textbf{\color{cmblu}83.5} &\textbf{\color{cmblu}90.2} &45.0 \\
        % \midrule
        % \multirow{3}{*}{$\cal{L}_{\rm{SCE}}$} 
        % &{\bf {\modelnameshort}}+$||\cdot||_1$  &82.3 &88.2 &44.9 \\
        % &{\bf {\modelnameshort}}+$||\cdot||_2$  &82.8 &88.2 &40.3 \\
        % &{\bf {\modelnameshort}}+${\rm{KL}}$  &\textbf{\color{cmblu}83.5} &89.1 &\textbf{\color{cmblu}49.1} \\
        % \midrule
        % \rowcolor{gray! 40} \multicolumn{2}{c}{\textbf{{\bf {\modelnameshort}}}}  &83.3 &89.2 &48.4 \\
        \multirow{3}{*}{$\cal{L}_{\rm{UN}}$} 
        &{\bf {\modelnameshort}}+ENT   &83.5  &90.4   &46.8 \\
        &{\bf {\modelnameshort}}+NRC   &83.9  &89.0   &47.5 \\
        &{\bf {\modelnameshort}}+TPDS  &\textbf{\color{cmblu}84.6} &\textbf{\color{cmblu}90.9} &45.0 \\
        \midrule
        \multirow{3}{*}{$\cal{L}_{\rm{SCE}}$} 
        &{\bf {\modelnameshort}}+$||\cdot||_1$  &83.4 &89.3 &44.9 \\
        &{\bf {\modelnameshort}}+$||\cdot||_2$  &83.8 &89.2 &40.3 \\
        &{\bf {\modelnameshort}}+${\rm{KL}}$    &84.5 &90.0 &\textbf{\color{cmblu}49.1} \\
        \midrule
        \rowcolor{gray! 40} \multicolumn{2}{c}{\textbf{{\bf {\modelnameshort}}}}  &84.3 &90.3 &48.4 \\
        \bottomrule
    \end{tabular}
    \label{tab:extend-exp}
\end{table}

\begin{table}[t]
    \centering
    \renewcommand\tabcolsep{1.5pt} 
    \renewcommand\arraystretch{0.9} 
    \scriptsize
    % \footnotesize
    \caption{Prompt initialization study~(\%).} 
        \begin{tabular}{l| l c c c}
            \toprule
            \multirow{3}{*}{\#} &\multirow{3}{*}{Initialization template}   &\multicolumn{2}{c}{Closed-set} &{SF-OODG} \\
            \cmidrule(lr){3-4} \cmidrule(lr){5-5}
            & &{\bf Office-Home} &{\bf VisDA} &{\bf IN-K} \\
            \midrule
            % 1 &'X [CLASS].' (\#X=4)         &82.6 &89.1 &48.4 \\
            % 2 &'X [CLASS].' (\#X=16)        &81.5 &88.5 &46.1 \\
            % \midrule
            % 3 &'There is a [CLASS].'            &83.6 &\textbf{\color{cmblu}89.2} &48.6  \\
            % 4 &'This is a photo of a [CLS].'  &\textbf{\color{cmblu}83.7} &89.0 &\textbf{\color{cmblu}48.8} \\
            % \midrule
            % 5 &'This is maybe a photo of a [CLS].'      &83.0  &89.1 &\textbf{\color{cmblu}48.8}  \\
            % 6 &'This is almost a photo of a [CLS].'     &\textbf{\color{cmblu}83.7} &\textbf{\color{cmblu}89.2}  &48.4 \\
            % 7 &'This is definitely a photo of a [CLS].' &83.5 &88.9  &48.1 \\
            % \midrule
            % 8 &'a picture of a [CLS].'  &83.5  &88.8 &48.6 \\
            % \rowcolor{gray! 40} 9 &'a photo of a [CLS].'  &83.3 &\textbf{\color{cmblu}89.2} &48.4 \\
            1 &'X [CLASS].' (\#X=4)                       &83.6 &90.1 &48.4 \\
            2 &'X [CLASS].' (\#X=16)                      &82.5 &89.6 &46.1 \\
            \midrule
            3 &'There is a [CLASS].'                      &84.6 &90.2 &48.6  \\
            4 &'This is a photo of a [CLASS].'            &84.7 &89.9 &\textbf{\color{cmblu}48.8} \\
            \midrule
            5 &'This is maybe a photo of a [CLASS].'      &83.9  &90.2 &\textbf{\color{cmblu}48.8}  \\
            6 &'This is almost a photo of a [CLASS].'     &\textbf{\color{cmblu}84.8} &\textbf{\color{cmblu}90.3}  &48.4 \\
            7 &'This is definitely a photo of a [CLASS].' &84.5 &90.0  &48.1 \\
            \midrule
            8 &'a picture of a [CLASS].'                  &84.6  &89.8 &48.6 \\
            \rowcolor{gray! 40} 9                         &'a photo of a [CLASS].'  &84.3 &\textbf{\color{cmblu}90.3} &48.4 \\
            \bottomrule
        \end{tabular} 
        \label{tab:ab_templ}
\end{table}
\begin{figure}[t]
    \begin{center}
        \subfigure{
            \includegraphics[width=0.6\linewidth]{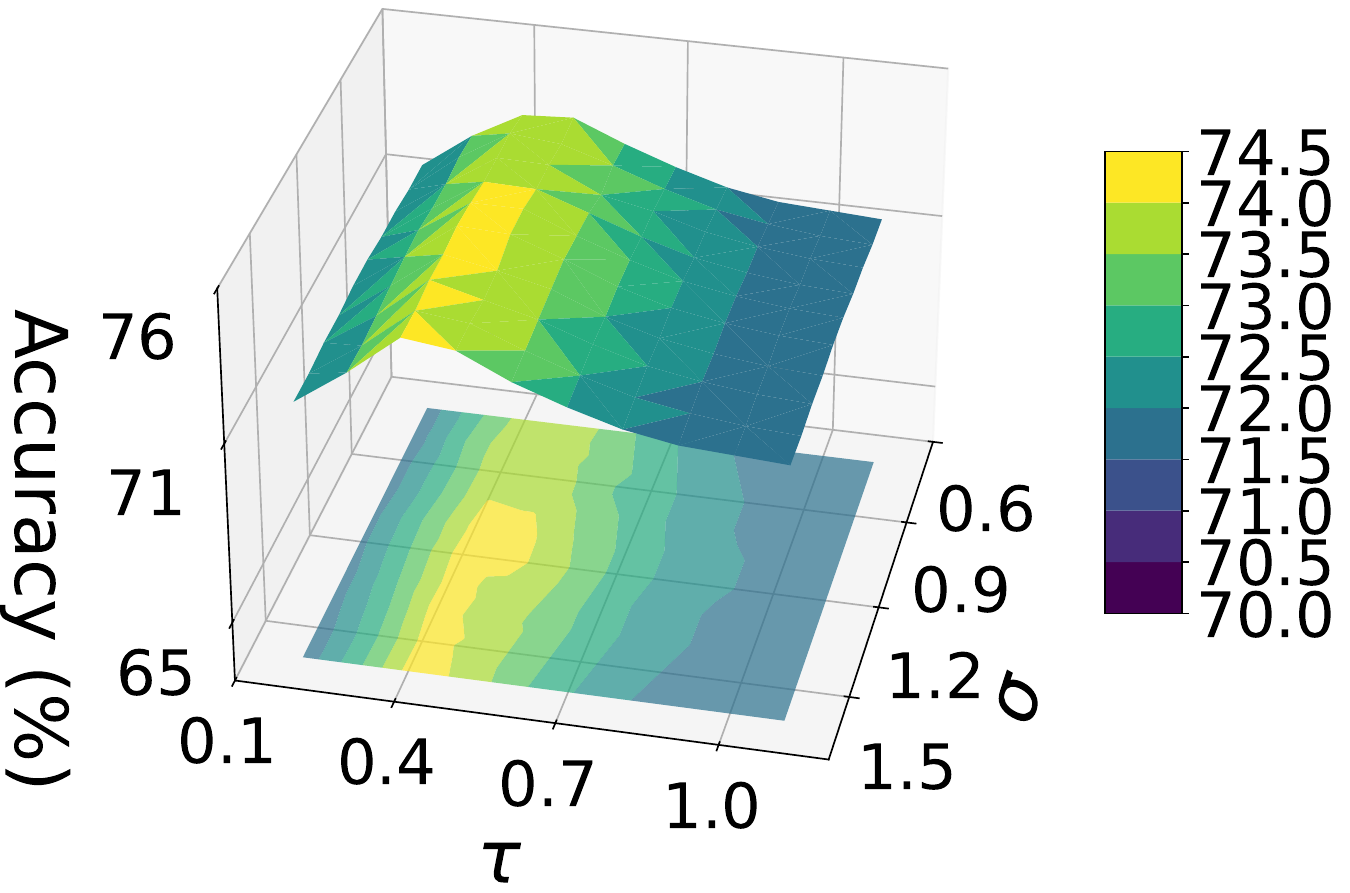}}~~
    \end{center}
    \setlength{\abovecaptionskip}{0cm}
    \caption{Parameter sensitiveness study ($\sigma$, $\tau$) on transfer task Ar$\to$Cl in {\bf Office-Home}.  
        }
    \label{fig:pars-sensi}
\end{figure}

\textbf{Effect of prompt initialization}.
In {\modelnameshort}, we employ prompt learning to encode the external causal factors \( S_e \). Here, we examine four different prompt initialization strategies (Tab.~\ref{tab:ab_templ}): (i) the conventional constant strategy (rows 1–2), (ii) the sentence strategy (rows 3–4), (iii) the \changedrr{uncertainty strategy} (rows 5–7), and (iv) an innovative approach of sentence initialization \changedrr{with different phrase} (rows 8–9), which considers the lack of precise supervision in SFDA.

Tab.~\ref{tab:ab_templ} shows the comparison results for these nine templates across both the Closed-set setting and SF-OODG. The results indicate that the different templates do not cause significant performance differences, suggesting that {\modelnameshort} is relatively insensitive to prompt selection. However, when comparing the constant strategy to the others, it becomes clear that initializing with semantic information is the more effective option, which aligns with our expectations.

\textbf{Parameters sensitiveness}.
Taking the transfer task Ar$\to$Cl in Office-Home for example, we evaluate the sensitivity of {\modelnameshort} by varying the hyper-parameters $0.2 \le \sigma \le 1.1$, $0.7 \le \tau \le 1.6$ at the step of 0.1. 
Fig.~\ref{fig:pars-sensi} indicates that both corresponding terms matter, whilst their sensitivity \changedrr{to} performance is not high.

\section{Conclusion} \label{sec:con}
We introduce a more challenging Unified SFDA problem that comprehensively addresses all specific scenarios within a unified framework.
To tackle this, we propose {\modelnameshort}, a novel approach grounded in causality, fundamentally differing from previous methods based on statistical dependence.
Specifically, we design a self-supervised information bottleneck to capture external causal elements, convert them into pseudo-labels, and use them to guide the discovery of internal causal factors.
Building on our theoretical insights, we implement this information bottleneck using a network-based variational mutual information approach.
Extensive experiments across various SFDA settings and SF-OODG validate the superiority of {\modelnameshort} over existing methods.

\subsubsection*{Acknowledgments}
This work is partially supported by the National Natural Science Foundation of China (NSFC) (62476169, 62206168, 62276048); the German Research Foundation and NSFC in project Crossmodal Learning under contract Sonderforschungsbereich Transregio 169, the Hamburg Landesforschungsf{\"o}rderungsprojekt Cross, NSFC (61773083); Horizon2020 RISE project STEP2DYNA (691154).

\ifCLASSOPTIONcaptionsoff
  \newpage
\fi

% trigger a \newpage just before the given reference
% number - used to balance the columns on the last page
% adjust value as needed - may need to be readjusted if
% the document is modified later
%\IEEEtriggeratref{8}
% The "triggered" command can be changed if desired:
%\IEEEtriggercmd{\enlargethispage{-5in}}

% references section

% can use a bibliography generated by BibTeX as a .bbl file
% BibTeX documentation can be easily obtained at:
% http://mirror.ctan.org/biblio/bibtex/contrib/doc/
% The IEEEtran BibTeX style support page is at:
% http://www.michaelshell.org/tex/ieeetran/bibtex/
%\bibliographystyle{IEEEtran}
% argument is your BibTeX string definitions and bibliography database(s)
%\bibliography{IEEEabrv,../bib/paper}
%
% <OR> manually copy in the resultant .bbl file
% set second argument of \begin to the number of references
% (used to reserve space for the reference number labels box)

%\begin{thebibliography}{1}
%\bibitem{IEEEhowto:kopka}
%H.~Kopka and P.~W. Daly, \emph{A Guide to \LaTeX}, 3rd~ed.\hskip 1em plus
%  0.5em minus 0.4em\relax Harlow, England: Addison-Wesley, 1999.
%\end{thebibliography}

{
    \bibliographystyle{IEEEtran}
    \normalem
    \bibliography{cas-refs-comm-simple}
}

\appendices
% \section{Theoretical Results}
\section{A Proof of Lemma 1}\label{sec-lema}
\noindent\textbf{Restatement of Lemma~1}
\textit{Given random variables $Z_1$, $X_1$, $Y_1$ where $X_1$, $Y_1$ satisfy a mapping $f_1\!:\! X_1 \!\mapsto \!Y_1$. 
When $f_1$ is compressed, i.e., the output's dimension is smaller than the input's, 
% then 
\begin{equation}
\small
    \label{eqn:pkpt-loss-upper}
    \begin{split}
        I\left( {{Z_1},X_1} \right) \geq I\left( {{Z_1},Y_1} \right)
    \end{split}
% \end{small}
\end{equation}
}
{\bf \textit{Proof.}}
Since mapping $f_1$ is compressed, the entropy of $X_1$, $Y_1$ have $H(X_1) \geq H(Y_1)$. As for 
$I\left( {{Z_1},X_1} \right)$, it has
\begin{equation}
\small
% \begin{small}
    \label{eqn:pkpt-loss-upper}
    \begin{split}
        I\left( {{Z_1},X_1} \right)
        &\geq I\left( {{Z_1},X_1} \right) - H\left(Z_1 | Y_1\right) - I\left(X_1,Y_1\right)\\
        &\geq \left[H\left(Z_1\right) + H\left(X_1\right)- H\left(Z_1,X_1\right)\right] - H\left(Z_1 | Y_1\right)\\ &~~~~~-\left[H\left(X_1\right)+H\left(Y_1\right)-H\left(X_1,Y_1\right)\right]\\
        % &= H\left(Z_1\right)- H\left(Z_1 | Y_1\right) - H\left(Z_1,X_1\right)-H\left(Y_1\right)+H\left(X_1,Y_1\right) \\
        &= H\left(Z_1\right)- H\left(Z_1 | Y_1\right) -H\left(Y_1\right)+H\left(X_1,Y_1\right) \\
        &\geq I\left( {{Z_1},Y_1} \right) -H\left(Y_1\right)+\max \left(H\left(X_1\right), H\left(Y_1\right)\right) \\
        &\geq I\left( {{Z_1},Y_1} \right) -H\left(Y_1\right)+ H\left(X_1\right)\\
        &\geq I\left( {{Z_1},Y_1} \right) -H\left(Y_1\right)+ H\left(Y_1\right)\\
        &=I\left( {{Z_1},Y_1} \right)
    \end{split}
% \end{small}
\end{equation}

\section{A Proof of Theorem 1} \label{sec-thm}
\noindent\textbf{Restatement of Theorem~1}
\textit{
Suppose there are five random variables, $Z$, $V$, $Z{'}$, $Y$ and $Y{'}$. Among them, $Z$ represents the target domain knowledge; $V$, $Z{'}$ and $Y$ express the input instances, predictions and intermediate features of ViL model, respectively; $Y{'}$ depicts a pseudo-label that has a learnable functional relationship $f_w$~($w$ is parameters) with $Z{'}$. When the function $f_w$ is reversible and uncompressed, we have this upper bound as follows. 
\begin{equation}
    % \begin{small}
    \small
    \begin{split}
    &\min_{\boldsymbol{v}_d} I\left( {Z,{Z{'}}} \right) - I\left( {{Z{'}},Y} \right)\\
    &\le \min_{\boldsymbol{v}_d,w} \underbrace {I\left( {Z,{Z{'}}} \right)}_{{T_1}} - \underbrace {I\left( {{Y{'}},Y} \right),{Y{'}} = f_{w}\left( {{Z{'}}} \right)}_{{T_2}}. 
    \end{split}
    % \end{small}
\end{equation}
}
{\bf \textit{Proof.}}
Since the function $f_w$ is reversible and uncompressed, the $Z{'}$ and $Y{'}$ satisfy bidirectional one-to-one mapping. In other words, the from $Y{'}$ to $Z{'}$, there is not information loss, formally $H\left(Y{'}\right)=H\left(Z{'}\right)$, termed condition A. 
Let $L=I\left( {Z,Z{'}} \right) - I\left( {Z{'},Y} \right)$, we have 
\begin{equation}
    \small
    \label{eqn:z-z-1}
    \begin{split}
        L &\leq I\left( {Z,Z{'}} \right) - I\left({Z{'},Y} \right) + I\left({Z{'},Y{'}} \right)\\
        &\leq I\left( {Z,Z{'}} \right) - \left[ H\left({Z^{'}|Y{'}} \right) - H\left({Z{'} | Y} \right)\right].
    \end{split}
\end{equation}
%改等号
The following discusses separately the two components in the second item of Eq.~\eqref{eqn:z-z-1} under condition A. 

\vspace{5pt}
\noindent(1) {\bf Proof} $H\left({Z{'}|Y{'}} \right)=H(Y^{'})$. 

Let $L_1 = H\left({Z{'}|Y{'}} \right)$, we have the relation as
\begin{equation}
    \small
    \label{eqn:z-z}
    \begin{split}
        L_1 &= H\left({Z{'}, Y{'}} \right) - H\left(Y{'} \right)\\
            &\leq  H\left(Z{'} \right) + H\left(Y{'} \right) - H\left(Y{'} \right) = H\left(Z{'} \right)
    \end{split}
\end{equation}
Also, we know $L_1 \geq H\left(Z{'}\right)$. Combing Eq.~\eqref{eqn:z-z}, we obtain $L_1 = H\left(Z{'}\right)= H\left(Y{'}\right)$ according to Sandwich Theorem and condition A. 

\vspace{5pt}
\noindent(2) {\bf Proof} $H\left({Z{'}|Y} \right)=H\left({Y{'}|Y} \right)$. 

Let $L_2 = H\left({Z{'} | Y} \right)$, $L_3 = H\left({Y{'} | Y} \right)$, here, we prove $L_2 = L_3$ with condition A. For $L_2$, it can be expressed as
\begin{equation}
\small
L_2 = H\left(Z{'}, Y \right) - H\left(Y \right),
\end{equation}
then we have  
\begin{equation}
    % \label{eqn:z-z}
    \small
    \begin{split}
        \max \left(H\left(Z{'}\right), H\left(Y \right)\right) - H\left(Y \right) &\leq L_2 \leq H\left(Z{'} \right)\\
        H\left(Z{'}\right) - H\left(Y \right) &\leq L_2 \leq H\left(Z{'} \right).
    \end{split}
\end{equation}
Similarly, $L_3$ can be expressed as  
\begin{equation}
    \small
    % \label{eqn:z-z}
    \begin{split}
        H\left(Y{'}\right) - H\left(Y \right) &\leq L_3 \leq H\left(Y{'} \right).
    \end{split}
\end{equation}
% Thus, 
\begin{equation}
    \small
    \begin{split}
        \rm{Thus},~H\left(Z{'}\right)\! - \!H\left(Y{'}\right) &\leq L_2\! - \!L_3 \leq H\left(Z{'} \right)\! - \!H\left(Y{'} \right).
    \end{split}
\end{equation}
Due to Sandwich Theorem and condition A, we have $L_2-L_3 = 0$, such that 
\begin{equation}
    \small
    \begin{split}
        L &\leq I\left( {Z,Z{'}} \right) - \left[ H\left(Y{'} \right) - H\left({Y{'} | Y} \right)\right] \\
        & = I\left( {Z,Z{'}} \right) - I\left( {Y{'},Y} \right).
    \end{split}
\end{equation}

\vfill

% You can push biographies down or up by placing
% a \vfill before or after them. The appropriate
% use of \vfill depends on what kind of text is
% on the last page and whether or not the columns
% are being equalized.

%\vfill

% Can be used to pull up biographies so that the bottom of the last one
% is flush with the other column.
%\enlargethispage{-5in}

% that's all folks
\end{document}